\documentclass[onecolumn,12pt]{IEEEtran}
\IEEEoverridecommandlockouts
\renewcommand{\baselinestretch}{1.25}

\usepackage{graphicx}
\ifCLASSINFOpdf
\else
\fi

\usepackage{amsmath,comment}

\usepackage{url}

\begin{document}
\title{Adversarial Learning in Statistical Classification: A Comprehensive Review of Defenses Against Attacks}

\author{David~J. Miller,
	Zhen~Xiang, 
        and~George~Kesidis
\thanks{The authors are with the School of EECS,
Pennsylvania State University,  University Park, PA, 16803, USA
email: \{djm25,gik2\}@psu.edu}
}


\maketitle

\begin{abstract}
With the wide deployment of machine learning (ML) based systems for a variety of applications including
medical, military, automotive, genomic, as well as multimedia and social networking, there is great potential for damage from 
{\it adversarial learning (AL) attacks}.  In this paper, we provide a
contemporary survey of AL, focused particularly on defenses against attacks on deep neural network classifiers.  After introducing relevant terminology and the goals and range of possible knowledge of both attackers and defenders, we survey recent work on test-time evasion (TTE), data poisoning (DP), backdoor DP, and reverse engineering (RE) attacks and particularly defenses against same.  In so doing, we distinguish
robust classification from anomaly detection (AD), unsupervised from supervised,
and statistical hypothesis-based defenses from ones that do not have an explicit null (no attack) hypothesis.  We also consider several scenarios for detecting backdoors.  We provide a technical assessment for reviewed works, including identifying any issues/limitations, required hyperparameters, needed computational complexity, as well as the performance measures evaluated and the obtained quality.  We then dig deeper, providing
novel insights that challenge conventional AL wisdom and that target unresolved issues, including: 1) robust
classification versus AD as a defense strategy; 2) 
the belief that attack success increases with attack {\it strength},  which 
ignores susceptibility to AD;
3) {\it small} perturbations for test-time evasion attacks: a fallacy or a requirement?;
4) validity of the {\it universal} assumption that a TTE attacker knows the ground-truth class for the example to be attacked; 
5) black, grey, or white box attacks as the standard for defense evaluation; 6) susceptibility of query-based RE to an AD defense.
We also discuss attacks on the privacy of training data.
We then present benchmark comparisons of several defenses against
TTE, RE, and backdoor DP attacks on images.  The paper
concludes with a discussion of continuing research directions, including the supreme challenge of detecting attacks whose goal is {\it not} to alter classification decisions, but rather simply to embed, without detection, ``fake news'' or other false content. 
\end{abstract}


%


\section{Introduction}
Machine learning (ML) based systems -- particularly, in recent years,  deep neural networks (DNNs) --  have found broad applications ranging from military, industrial, medical, multimedia/Web, and scientific (including genomics) to even the political, social science, and legal arenas. 
As their integration into the modern world's infrastructure continues to grow,
they become ever more enticing targets for {\it adversaries}, including individual hackers, criminal
organizations, as well as government intelligence services, which may seek to ``break'' them.  Thus, {\it adversarial learning} (AL),
the problem of devising attacks against ML systems as well as defenses against such attacks \cite{Tygar},\cite{Biggio_wild}, has become
a popular topic over the past decade.  We exclude from consideration here methods such as Generative Adversarial Networks (GANs) \cite{gans}, where the ``adversary'' is just a component of a game-theoretic ML
framework, under complete control of the ML designer -- here the focus is on adversaries which
seek to disrupt operation of the ML system in some way.  Researchers have devised attacks against various
types of ML systems, including supervised classifiers and regression/prediction \cite{Szegedy_seminal}, as well as 
unsupervised clustering and density estimation \cite{Biggio_unsup}.  The main focus of this paper is on contemporary attacks against supervised classifiers.  Moreover, since in many applications DNNs achieve
state-of-the-art performance, the recent focus has been on attacking DNNs.  Thus, such attacks,
and especially defenses against them, is central to this survey.  The goal is to provide
a survey as comprehensive and as technically probing as possible, given space limitations.

\noindent
{\it Brief Review of Statistical Classification:}
\noindent
A classifier is a function $C(\underline{x}) \in {\cal K} \equiv \{\omega_1,\ldots,\omega_K\}$ that acts on an input, fixed-dimensional feature vector 
$\underline{x}$, whose entries could be numerical, categorical, or ordinal, and maps the feature vector to
one of $K$ categories defined for the given application domain. 
For example, for the MNIST image data set \cite{mnist}, the ten
classes are the digits $\omega_1=$`0' through $\omega_{10}=$`9'. The feature vector 
(sometimes also called the ``pattern'') could be {\it raw} ({\it e.g.} a speech waveform or a scanned digital image) or it could be {\it derived}, based on some front-end feature extraction ({\it e.g.}, linear prediction or cepstral coefficients for speech \cite{Rabiner}, a bag-of-words representation for a text document, {\it e.g.} \cite{Nigam2000}, or principal component representation of an image).  Often, the classification mapping is implemented by a ``winner-take-all'' (WTA) rule \cite{Duda} applied to a collection of (learned) discriminant functions, one per class. WTAs are used both by neural network based classifiers, including DNNs \cite{Goodfellow}, as well as linear and generalized linear classifiers ({\it e.g.}, support vector machines (SVMs) \cite{Vapnik}).  Another widely used classifier structure is a decision tree \cite{Breiman_book}.
For a review of various classification methods, see \cite{Duda}. 

DNNs perform nonlinear processing via multiple layers of computational neurons.  DNNs can be feedforward (the focus here) or recurrent.  Neurons can use one of several activation functions
({\it e.g.}, sigmoid, rectified linear unit (ReLU)).  Fully connected weights may connect one layer to the next; alternatively, a convolution layer of weights may be used. DNNs are not in fact new -- the use of convolutional layers to extract key features (feature maps) while being model-parsimonious (via weight tying) is both quite an old idea \cite{LeCun} and a mainstay of contemporary DNNs.
Modern DNNs also often use {\it e.g.} neuron dropout, batch normalization, data augmentation, and stochastic gradient descent for learning \cite{Goodfellow}.

The decision rule of a classifier divides the feature space into mutually exclusive and collectively exhaustive regions, one per class. Neighboring classes are those which share a decision boundary (the boundary between two classes, if it exists, is the locus of points for which the two classes have the same (largest) discriminant function value). The classifier is trained, based on an available labeled training set
$${\cal X}_{\rm train} = \{(\underline{x}_i,c_i): i\in\{1,\ldots,N_{\rm train}\}, c_i \in {\cal K}\},$$
to minimize a (supervised) loss function measured over the training samples.  The goal is to learn a decision rule that generalizes to make accurate decisions on test samples, unseen during training, drawn according to the true joint density
function $f_{\underline{X},C}(\cdot)$, which is generally unknown.
The ultimate attainable performance for a given classification problem is the Bayes error rate, which 
depends on this (unknown) joint density \cite{Duda}.  The classifier's generalization error is often estimated using a separate, finite labeled test set (held out from training).  Often, the labeled data resource is split into exclusive training and test subsets. Cross validation methods are also used to estimate generalization error.  One may also take some of the training data and use it as a validation set for choosing {\it hyperparameters} of the classifier that cannot be easily chosen as part of the standard classifier
learning algorithm.  For example, for an SVM, parameters of the kernel (even the choice of kernel) and the degree of slackness on margin violations are hyperparameters that may need to be chosen.  For a DNN, the number of layers, the sizes of the layers, the layer type (fully connected, convolutional, or maxpooling), the choice of activation function, whether a form of normalization is applied to a given layer, the stopping rule for learning, and the neuron ``dropout" rate \cite{dropout} could all be considered hyperparameters, which makes for a large hyperparameter space. 

\noindent
{\it Terminology for and Types of Adversarial Learning Attacks:}
Early seminal work introduced useful terminology to distinguish various types of AL attacks and their distinct
objectives at a high level \cite{Tygar}.  {\it Causative} attacks alter the learned classification model.  {\it Exploratory}
attacks do not alter the learned model but seek to learn information either about the classifier/decision rule
or about the data samples on which it was trained (which may contain private or sensitive information).
{\it Targeted} causative attacks focus on ensuring the classifier assigns either a particular 
subset of data samples (even a single sample) or a particular region of feature space to a chosen (target) class.
{\it Indiscriminate} causative attacks, on the other hand, simply seek to induce decision changes without ensuring a particular 
target class is assigned.  Moreover, {\it availability} attacks seek to make a classifier unusable by degrading
its accuracy to an unacceptably low level.
There are three fundamental types of AL attacks on classifiers, which we next describe.

\noindent
{\bf Data Poisoning (DP):}  These are causative attacks wherein the attacker has the ability to introduce ``poisoned''
samples into the training (and/or validation and/or test) sets.  The poisoned samples could either be 
joint feature density ``typical'' examples from the domain but which are {\it mislabeled} (either in a targeted
or indiscriminate fashion) or examples that are not typical of the domain.  For example, if the image domain is 
categories of vehicles, a bird image would be atypical.  

Until recently, most DP attacks were {\it availability} attacks, seeking to degrade the learned classifier's accuracy, {\it e.g.} \cite{Tygar11},\cite{Xiao15},\cite{MLSP17}.  However, there is also strong recent interest in {\it backdoor} DP attacks 
\cite{Song},\cite{Madry-NIPS18},\cite{BadNet},\cite{Haoti},\cite{Trojan},\cite{MLSP19-backdoor},\cite{backdoor-perceptible}, {\it i.e.} targeted
attacks which seek to have the classifier learn a ``backdoor'' pattern embedded into the poisoned samples.
The backdoor pattern could be an imperceptible (random) watermark-like pattern or something
perceptible but innocuous -- {\it e.g.}, the presence of glasses on a face \cite{Song}.
Backdoor attacks seek for the classifier to learn to classify to the target class when the backdoor pattern is present but otherwise not to alter its decisionmaking -- degradation in accuracy in the absence of the backdoor pattern could destroy the classifier's usability and could also be the basis for detecting that the classifier has been DP-attacked.

In some works, it is assumed that initially there is a clean 
(free of poisoned samples) 
training set, but that it is subsequently altered by additional data collection, by on-line learning, or by the actions of an adversarial insider.
Under this scenario, for availability DP attacks, one can detect poisoned samples by discerning that their use
in learning degrades classification accuracy (on a clean validation data subset) \cite{Roni} relative to just use of the clean data. 
A more challenging
DP scenario for attack detection is the {\it embedded} scenario, where one cannot assume the training
data set is initially clean
and where there is no available means (time stamps, data provenance, etc.) for identifying a subset of samples guaranteed to be free of poisoning.  

\noindent
{\bf Reverse Engineering:}  These are exploratory attacks that involve querying (probing) a classifier either 
to learn its decision rule or to learn something about the data set on which it was trained (an attack on data privacy).  
Querying creates a training set for the attacker, allowing him/her to learn a surrogate of the true classifier.
Several motivations have been given for reverse-engineering a classifier's decision rule.  In \cite{Reiter},
the goal was to avoid having to pay for use of an on-line ML service that makes decisions, given a (paying) user's queries.  A more concerning attacker objective is to learn a good surrogate classifier so that the
attacker can identify the classifier's ``vulnerabilities'' -- in particular, knowledge of the classifier allows
the attacker to craft {\it test-time evasion attacks}, described next.

\noindent
{\bf Test-Time Evasion:} 
This fundamental attack on classifiers is one wherein test (operational) examples are perturbed, either human-imperceptibly and/or so that they are not easily machine-detected, but in such a way that the classifier's decision is {\it changed} (and now disagrees with a consensus human decision), 
{\it e.g.} 
\cite{Biggio_seminal,Szegedy_seminal,Papernot,Goodfellow,CW}.  
TTE attacks may be either targeted or indiscriminate, although, as with backdoor attacks, they are most strategic
if they are {\it targeted}, perturbing samples from a source category $c_s$ so that they are classified to 
a particular target (destination) category, $c_d$.  In order to ensure such an attack is successful, the attacker needs to know the true class label of the pattern to be attacked. Crafting TTE attacks involves perturbing a source pattern until it moves from the decision region for one category across the decision boundary into the region for another ({\it e.g.}, targeted) category.
Such attacks, creating {\it adversarial
examples} \cite{Szegedy_seminal}, may {\it e.g.} cause an autonomous vehicle to fail to recognize a road sign, cause an automated system to falsely target a civilian vehicle, or grant image or audio authentication access to a building, machine, or to restricted information.
Great interest in AL was generated by \cite{Szegedy_seminal}, which showed DNNs are quite susceptible to TTEs.
A TTE attack may be on physical objects in the real world ({\it e.g.}, altering a road sign, camouflaging a vehicle, or darkening a room to make it more difficult to identify objects of interest).  Alternatively, they may be alterations of data objects that have either already been digitally captured (digital images, voice files) or those which are {\it natively} digital, {\it e.g.}, emails, documents, or computer programs.  If a TTE attack is enabled by an RE attack, which learns a surrogate classifier (rather than assuming the TTE attacker possesses perfect knowledge of the actual classifier), an important property is {\it transferability} of the TTE attack \cite{Papernot3}: does a perturbation of a pattern that induces a targeted or indiscriminate change in the surrogate classifier's decision also induce such a change in the actual classifier's decision?

\noindent
{\it Knowledge and Assumptions of the Attacker and the Defender:}
In publications involving the aforementioned attack types, various assumptions, either explicit or implicit, are made about the
knowledge and resources of the attacker as well as those of the {\it defender}, {\it i.e.} the system consisting of the classifier plus
any defense which it mounts to defeat potential attacks. \cite{Biggio_KDE} has advocated making the assumptions and goals of the attacker explicit in
 AL work.  We further advocate making explicit the {\it defender's} goals and assumptions.
 In particular, an AL researcher must compartmentalize his/her knowledge in devising a defense, in
 defining an attack which will be evaluated against the defense, and in conducting experiments -- explicitly stating knowledge and assumptions of both sides helps a reader to verify no hidden assumptions are being made in devising a defense, which help it to defeat an attack, or vice versa; also, it helps in assessing whether a paper's experiments truly reflect the knowledge and assumptions being made on both sides.
Again, some useful terminology has been introduced \cite{Tygar}.  First, an attacker
or defender may be either {\it proactive} or {\it reactive}.  A proactive defense is one mounted without specific knowledge of a potential
attacker's strategies, objectives, or knowledge.  The defender's challenge here is formidable, since he/she is seeking to protect the
classifier against any and all potential attacks, but not necessarily insurmountable.  Against TTE attacks, proactive defenses include some
{\it robust classification} strategies, which seek to correctly classify attacked patterns, as well as unsupervised (without attack examples) {\it anomaly detectors}, relying
on attacked patterns exhibiting detectable atypicality relative to clean patterns.  Similarly, robust classification and detection have
been proposed as proactive strategies against DP attacks, including recent detection of backdoor
attacks \cite{Madry-NIPS18},\cite{BChen18},\cite{MLSP19-backdoor}.  
In \cite{backdoor-perceptible,TrojAI,NC}, notably, backdoors are detected {\it without} access to the poisoned training set.
A {\it reactive} attack is one which exploits knowledge of a mounted defense.
Quite a few papers have shown that reactive attacks can defeat proactive defenses by specifically targeting them, e.g. \cite{CW,Zhang-blindspot}.  This setting is implicitly game-theoretic, with the (sequential) advantage to the attacker.  Presumably, with knowledge
of the attack, the defender could alter his/her strategy to defeat 
it.  For example, \cite{Metzen} developed a {\it supervised}
detector of TTE attacks.  Unfortunately, while successful in detecting assumed known attacks, this method fares poorly if the attacker
uses an {\it unknown} strategy \cite{Metzen,Wagner17}.   

A wide range of assumptions about the attacker's knowledge have been made in the AL literature.  Early DP work \cite{Lowd} targeting naive Bayes
email-spam filters showed that: i) TTE attacks could be mounted by adding ``good'' words, typically appearing in ham (non-spam) emails, to spam emails;
ii) likewise, effective DP attacks are created by sending ``ham-like'' emails from a black-listed IP address, which are treated
as labeled spam instances and used to 
on-line  refine (in this case, corrupt) the naive Bayes spam filter.  

These attacks merely assumed some knowledge of ``good'' and ``bad'' words.  Subsequent works have assumed much greater attacker 
knowledge, as well as capabilities and resources.  In particular, for both DPs and TTEs, the attacker may possess knowledge of:
i) the feature space; ii) the classifier type ({\it e.g.}, SVM or DNN); iii) the classifier's learning algorithm; iv) any hyperparameters associated with classifier training; v) the training set used by the classifier.
In the TTE case, if the attacker knows all of the above, he/she has perfect knowledge of the actual classifier {\it and} a rich resource of legitimate supervised examples from the domain.  In the DP case,
the attacker can formulate an optimization problem seeking to maximally degrade accuracy with the fewest possible poisoned samples.
In \cite{KLiu18}, it is even assumed that classifier learning is outsourced to a third party who is also the DP attacker.  In this highly vulnerable scenario, the attacker has great facility either to degrade accuracy in a controlled way
or to embed backdoors into the learned classifier.

Full knowledge (of the classifier plus any mounted defense on top, such as an anomaly detector) has been referred to as a {\it white box} assumption\cite{Wagner17}.  At the other extreme, no knowledge of the classifier
(but some ability to query it) is a {\it black box} assumption.  A grey box attack is somewhere in between -- in \cite{Wagner17},
assumptions i) - iv) above are made, but, rather than possessing the full training set, the TTE attacker's surrogate classifier is trained using other
data (still representative of the domain, but perhaps smaller than the training set possessed by the learner/defender).
Arguing that {\it security by obscurity} is unreliable \cite{Biggio_wild}, it is plausible an attacker may possess full (white box) knowledge
of the classifier.  However, we will argue in Section IV that it seems less plausible the attacker will also possess full knowledge of any {\it detector} deployed along with the classifier.  

Some attack works also make {\it implicit} assumptions.  \cite{Papernot} generates TTE attacks both with noticeable artifacts
(on the MNIST image digits domain) and which are in some cases class-ambiguous to a human being (see Figure 4).  These attacks are deemed successful in \cite{Papernot} but
human monitoring would deem these images suspicious \cite{MLSP18-ADA} has also shown that a very simple detector has strong capability to detect this attack.).
Moreover, if there is substantial decision uncertainty, the classifier can use a {\it rejection} (or ``don't know") option.  This
is not allowed in \cite{Papernot}.  Rejection is essential when the problem is not classification
per se, but {\it authentication} ({\em e.g.,} for accessing a building or restricted information).  Many TTE attack works also do not consider
existing, conventional defenses that may be deployed.  For example, for a ``man-in-the-middle'' scenario where the pattern is transmitted to the
classifier but intercepted on the way by the TTE attacker, use of 
standard encryption techniques
would defeat the attack \cite{kurose}. 
Likewise, in \cite{Wagner16}, a 
TTE attack on voice systems could potentially be defeated by SIRI's 
existing built-in speaker-recognition system\footnote{Though a
speaker-recognition system can be overcome by  existing
voice cloning technology (especially under a  white box
scenario where the attacker has samples of the authorized
party's voice), standard techniques of limited privilege
can be applied. For example, 
Siri cannot enter data into Safari, and 
even if it could, there is  standard
two-factor authentication 
to prevent unauthorized access to a private web site.  Operationally,
such security measures may require significant
overhead and could  be instituted only {\em in response}
to detected attacks, {\it cf.} Section \ref{sec:AD-TTEs} where
detection of TTE attacks is discussed in detail.}.
Moreover, all TTE attacks assume the attacker knows the true class of the example to be altered.  In Section IV we show this assumption is {\it not} valid in one practical setting (a man-in-the-middle).

Another important characterization of an attack is its {\it strength}.  For TTEs, attack strength is quantified by some measure of the
amount of perturbation from the original, clean pattern.  For general DP attacks, attack strength
is two-dimensional: the amount of perturbation per poisoned pattern and the fraction of the training
set that is poisoned.
For backdoor attacks, the fraction of poisoned patterns is one measure; moreover, if a watermark-like pattern is embedded, the signal-to-noise ratio (treating the backdoor pattern as noise) is a second
attack strength measure.  While one might expect that the attacker's success rate will increase with its strength, a stronger attack may 
also be more {\it detectable} (see Section IV).

\noindent
{\it Performance Measures:}  For general DP attacks, the attack success can be quantified by the amount of induced degradation in the
classifier's (test set) correct decision rate.  For backdoor attacks it can be quantified by the rate of classifying backdoor patterns to the target
class {\it and} by any increased degradation in the classifier's accuracy.  However, if a DP attack {\it detector} is in play, then the
true positive rate (TPR) and false positive rate (FPR) of the detector are of great interest.  All of these quantities could vary as a 
function of attack strength.

For TTE attacks, there are multiple, vital performance criteria, some of which have not been consistently considered in past works.
First, there is the attack success rate in inducing (possibly targeted) decision changes {\it and} in avoiding detection, if a detector
is deployed.  If a robust classifier is used, there is the correct decision rate of the classifier in the presence of TTE attacks.
There is also the correct decision rate of the classifier in the {\it absence} of TTE attacks (A robust classifier may degrade 
accuracy in the absence of an attack \cite{Madry-robust,Li_ICCV}.  
Also, a detector may {\it reject} some attack samples, as well as some non-attacked test samples,
which can also affect the classifier's accuracy.).  If a detector is deployed, there is the true attack detection rate (TPR) and
false positive rate (FPR), and the total area under the (FPR, TPR) curve (the ROC AUC).  All these quantities may vary as a function of attack strength.  

Finally, there is the computational workload of the attacker in creating DP and TTE attacks, and of the robust classifier and/or detector, in
defending against them.  For example, in \cite{MLSP18-ADA}, the workload of the CW attack was found to be much higher than that of the AD defense.

\noindent
{\it Classification Domains:} Most AL work has focused on image classification.  Two comments
are in order here.  First, even though they have been applied to images, some attacks, as well
as some defenses, are actually more general-purpose, applicable {\it e.g.} whenever data samples are represented as vectors of continuous-valued features.  For example, methods applicable to images may also be applicable to speech waveforms (or associated derived feature vectors), to gene microarray expression profiles, and even to text documents ({\it e.g.}, if documents are represented as continuous-valued feature vectors via latent semantic indexing, principal component analysis, or word2vec mappings).  Thus, to emphasize the potential general applicability of a particular attack or defense, we will refer to its operation on a {\it pattern} (feature vector), rather than
on an image.  Second, we will also discuss some works that focus on domains other than images.

\noindent
{\it The rest of the paper is organized as follows.}
In Section \ref{sec:attack}, we review AL attacks.
In Section \ref{sec:defense}, we review recent work on AL defenses.
In Section \ref{sec:dd},  we offer some deeper perspectives, particularly
regarding TTE attacks.
In Section \ref{sec:expt}, we provide some experimental studies.
Finally, in Section \ref{sec:fw}, we discuss future work.

\section{Review of AL Attacks}\label{sec:attack}
\subsection{TTE Attacks}
Early work considered the spam email domain and relatively simple classifiers
such as naive Bayes classifiers, using ``good word'' attacks \cite{Lowd}.
Subsequently, considering a two-class problem, 
\cite{Biggio_seminal}
proposed a constrained objective function for the TTE attacker, seeking (in the two-class case)
to minimize the discriminant function output while constraining the amount of perturbation from
the original, clean pattern.  They demonstrated their approach for 
SVM classification of images and PDF files and also showed how it could be 
extended to neural network classifiers.
\cite{Szegedy_seminal} considered DNNs, posed a related attacker optimization objective, 
seeking imperceptible image perturbations that induce misclassifications, showed that this approach
is highly successful in creating adversarial examples starting from any (or at any rate, most) legitimate images, and also
showed that these adversarial examples seem to {\it transfer} well (remaining as adversarial examples
for other networks with different architectures and for networks trained on disjoint training sets). 
There are much older optimization approaches related to \cite{Biggio_seminal} and \cite{Szegedy_seminal} for finding decision boundaries in neural networks,
 e.g. \cite{Hwang97}, but which were not specifically applied to create adversarial test-time attacks.

A related method, proposed in \cite{Goodfellow}, the fast gradient sign method (FGSM), was
motivated primarily by the goal of creating adversarial examples that could be added to
the learner's training set (data augmentation)
and thus used to increase robustness of the learned classifier.
However, FGSM has also been frequently used as a TTE attack in works assessing TTE defenses.

While
\cite{Biggio_seminal,Szegedy_seminal,Goodfellow}, are ``global'' methods,
potentially altering all of a pattern's features ({\it e.g.}, all image pixels' values) but while constraining the overall distortion
of the pattern perturbation, \cite{Papernot}
proposed a TTE attack that restricts the number of modified features (pixels). 
However, this requires large changes on the modified features 
in order to
induce a change to the classifier's decision.
\cite{Papernot} cycles over features (pixels), one at a time, at each step modifying the pixel
estimated to effect the most progress toward the classifier's decision boundary.  This procedure is applied either until a successful decision change is induced or until the modified pixel
budget is reached.

An improvement over prior TTE attack works is achieved by ``CW'' 
\cite{CW,Wagner17}. Here, the attacker's objective function, to be minimized in crafting perturbations, 
focuses in a {\it discriminative} fashion \cite{Juang} on {\it two classes}: the target class ($t$)
and the class other than the target class with the largest discriminant function value, which
could be the true class of the original, clean pattern.  Let $F_j(\underline{x})$
denote class $j$'s discriminant function value for the original pattern, with the classifier
using the WTA rule $C(\underline{x}) = {\rm arg max}_j F_j(\underline{x})$.  
Then, \cite{CW} essentially forms the attacker's optimization problem as:
\begin{eqnarray}
\label{cw}
{\rm min}_{\underline{x}'} (||\underline{x}' - \underline{x}||_p + c
\cdot {\rm max}\{{\rm max}_{j\neq t}F_j(\underline{x}') - F_t(\underline{x}'),-k\}).
\end{eqnarray}
We make the following comments:
\begin{enumerate}
\item Minimizing (\ref{cw}) seeks the perturbed pattern that, while constrained in $p$-norm
distance to $\underline{x}$, maximizes the discriminant
function difference between that of the target class and that of the nearest competitor class, {\it i.e.} it maximizes {\it confidence} in decisionmaking to class $t$.  However, very high confidence misclassifications 
are not necessarily sought.  Thus, the hyperparameter $k$ controls the maximum allowed 
confidence in the decision.  While the authors show in \cite{CW} that increasing $k$ increases attack
{\it transferability}, they do not suggest how to choose $k$ in the case
where there is a detector in play which is (black box) unknown to the attacker -- it is unclear what level of decision ``confidence'' will be consistent with being detection-evasive.
\item The hyperparameter $c$ controls the cost tradeoff between inducing a misclassification and the amount of pattern perturbation.  If $c$ is too small, the relative penalty on the size of perturbations is too large to allow decision alteration.  If $c$ is too large, misclassifications will be induced but the perturbations may be large (and hence perceptible and/or
machine-detectable).  This hyperparameter is chosen via a binary search procedure, seeking the smallest
value inducing misclassifications.
\item The strongest version of the CW attack, observed experimentally \cite{Wagner17},\cite{MLSP18-ADA},  uses the $p=2$ norm.
\item \cite{MLSP18-ADA} also extends the attack to defeat both a classifier and an accompanying detector, {\it i.e.} they proposed a {\it ``fully'' white box} attack, exploiting full knowledge
of both the classifier and detector.  However, as will be discussed later, \cite{Wagner17} showed that even with just full knowledge of the classifier (white box with respect to the classifier, black box with respect to the detector), their attack defeats many TTE defense schemes, including some AD defenses.
\item While \cite{Wagner17} gives an extensive benchmark experimental study,
they do not consistently report FPRs for the detection methods they evaluated -- {\it e.g.}, consider the evaluation of \cite{AD} in the ''black box'' case in \cite{Wagner17}. 
\item \cite{Wagner17} also asserts that as a defender ``one must also show that a adversary 
aware of the defense can not generate attacks that evade detection''.  We refer to this
as the {\it strong white box} requirement, and discuss this further in 
Section IV.
\end{enumerate}

There is also
intriguing work on {\it universal} TTE attacks on DNNs, {\it i.e.} a {\it single} perturbation vector that induces
classification errors when added to {\it most} test patterns for a given domain \cite{Dezfooli}.  This work is important because it is suggestive of the fragility
of DNN-based classification. (It will also inspire a backdoor detection approach \cite{TrojAI}.)  However, a {\it single} perturbation vector required to induce errors
on nearly {\it all} test patterns must be ``larger'' than a customized perturbation designed to successfully misclassify
a {\it single} (particular) test pattern. 
Thus,
images perturbed by a universal attack should be more easily detected than those perturbed by the pattern-customized attacks in,
{\it e.g.},
\cite{Goodfellow,Papernot,CW,MLSP18-ADA}.  

Although the previous works could in principle be applied to any classification domain working in some (pattern) space, they all focused experimentally on images.
There has been work on other domains, including voice \cite{Wagner16,Wagner18} 
and music 
\cite{Larsen15a,Larsen15b}. \cite{Wagner16} developed an iterative procedure, with a human attacker in the loop, to generate voice commands that are recognized by a speech recognition system and yet are not human-understandable. \cite{Larsen15a,Larsen15b} considered music genre classification.
In \cite{Larsen15a}, the authors showed that changes to the tempo 
degrade the accuracy of a DNN   
but do not affect a human
being's capability to classify the musical genre.
This suggests
that the DNN is not using the same ``reasoning'' as a human being in performing music classification.  In \cite{Larsen15b}, the authors modify the TTE attack approach in \cite{Szegedy_seminal} to be suitable for the music domain.  This would be straightforward, except that the classification is performed on individual ``frames'' that are temporally overlapping. 
Thus, in constructing a TTE attack for a given frame, one should ensure consistency with all frames with which the given frame overlaps.  This problem is addressed in \cite{Larsen15b} using an iterative projection algorithm.   
A good reference more exhaustively covering the variety of TTE attacks on images that have been proposed, as well as papers that try (though with no closure to date) to explain why DNNs are susceptible to TTEs, is \cite{IEEE_access}.

\subsection{DP Attacks}

\begin{figure}[h]
\centering
\includegraphics[width=\columnwidth]{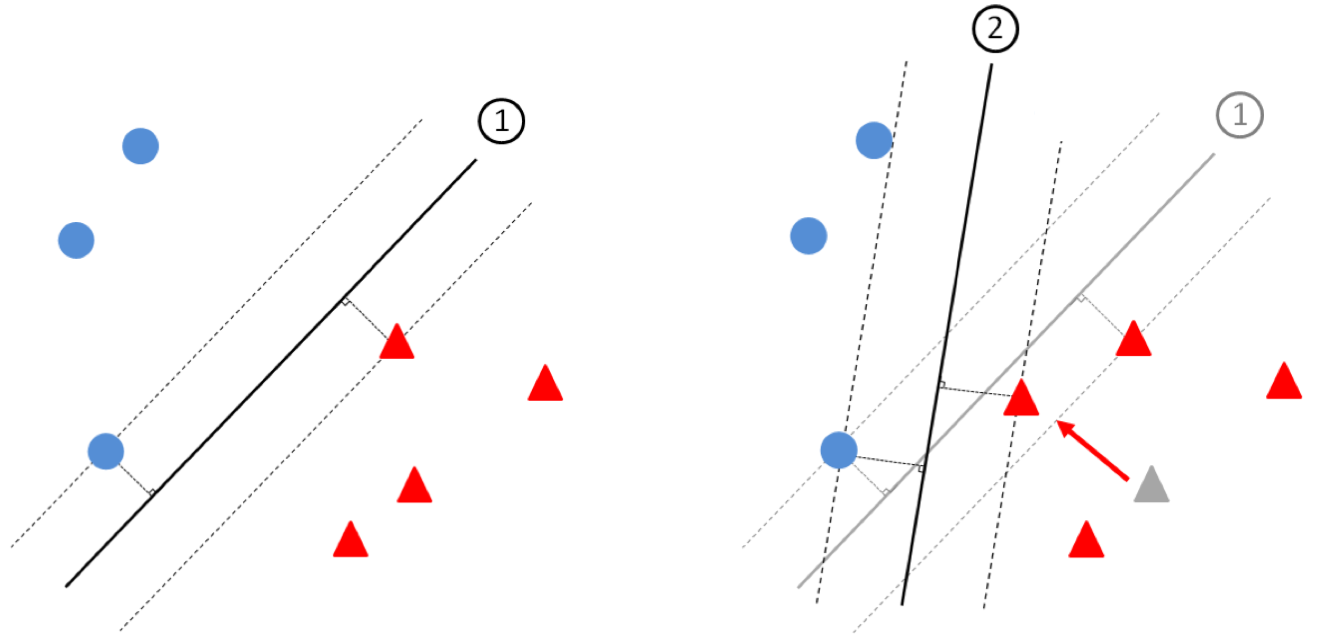}
\caption{Linear SVM classifier decision boundary for a two-class dataset with support vectors and classification margins indicated (left). Decision boundary is significantly impacted in this example if just one training sample is changed, even when that sample's class label does not change (right).}\label{fig:svm}
\end{figure}

As noted, early work on DP attacks addressed spam email and did not require the attacker to know very much \cite{Tygar}.
More recent availability DP attacks require greater knowledge of the system under attack.
\cite{Biggio_svm} showed that significant degradation in an SVM's accuracy could be achieved with the
addition of just one poisoned training sample -- on MNIST the error rate increased from
the 2-5\% range to the 15-20\% range.  To achieve this, the attacker exploits knowledge
of the training set, a validation set, the SVM learning algorithm, and its hyperparameters.
The authors define the attacker's objective function as classifier error rate on the validation set, as a function of the poisoned sample's location (and class label). 
This loss is maximized under the constraint that the support vector, non-support vector subsets
of the training set are not altered by addition of the poisoned sample.  Thus, \cite{Biggio_svm}
adds a single, new (adversarially labeled) support vector to the SVM solution.  The constraint
is met by performing gradient descent carefully, with a small step size.  One could expect
that even more classifier degradation could be achieved if the support vectors were allowed
to {\it change} through the addition of the poisoned sample\footnote{On the other hand, if the attacker does {\it not} know the value of the margin slackness hyperparameter, he/she cannot ensure the poisoned sample will be a support vector; in such case, many more poisoned samples may be needed, in order to significantly degrade SVM accuracy.}.  However, this would also entail a more complex optimization procedure.
An illustrative example of a DP attack on SVMs is shown in 
Figure \ref{fig:svm}.

While SVMs (which rely on a ``support vector'' subset of the training set to define the linear discriminant function) can unsurprisingly be fragile in the presence of availability DP attacks, there is little prior work
investigating such attacks against DNNs.  One reason may be computational complexity -- 
{\it e.g.}, one could in some way alternate gradient descent optimization in weight space (
minimizing the loss function, {\it i.e.} the defender/learner's problem) and
gradient {\it ascent} in pattern space (maximizing the loss function, {\it i.e.} the attacker's problem), to find a set of poison patterns that maximally degrade the learned DNN's accuracy.
However, such a procedure would be quite computationally heavy.  Moreover, DNNs should be less fragile to data poisoning than SVMs, whose solutions explicitly depend on (a subset of) individual
training points -- to degrade accuracy sufficiently a larger fraction of poisoned samples may be needed for DNNs.  For ``big data domains'', with {\it e.g.} one million training samples, even  10\% data poisoning means optimizing 100,000 poisoned sample locations and labels.

On the other hand, DNNs appear to be quite vulnerable to {\it backdoor} DP attacks, as
demonstrated in several recent works \cite{Song,Haoti}.   
The backdoor pattern could be an imperceptible ({\it e.g.} random-looking) watermark-like pattern or something
perceptible but innocuous -- {\it e.g.}, the presence of glasses on a face \cite{Song}, a plausible object in the background
in an image scene (such as a tree or a bird in the sky), or a noise-like audio background pattern in the case of speech classification.  One very attractive aspect of these attacks is that they may
require {\it no} knowledge of the classifier -- the attacker simply needs i) legitimate examples from the domain, into which it embeds the backdoor pattern; ii) the ability to poison the training set, with these samples labeled to the target (backdoor) class; iii) perhaps knowledge of the training set size, to know how many poisoned samples may be needed.
On the other hand, if the attacker possesses knowledge of the classifier, its training set, and its
learning algorithm, he/she can optimize the backdoor pattern in order to ensure: 1) the backdoor is well-learned; 2) classifier accuracy is not compromised; and 3) 1) and 2) are accomplished
with the least amount of attack strength (and hence the least attack-detectability).
An attempt at such an approach is given in \cite{Haoti}. 

\subsection{RE Attacks}
This is an important attack type, but an emergent one, {\it i.e.} this is not yet a mature research
area.
In \cite{Reiter}, the
authors consider black box ML services, offered by companies such as
Google, where, for a given (presumably big data, big model) domain, a user pays for class
decisions on individual
samples (queries) submitted to the ML service.  \cite{Reiter} demonstrates that, with a relatively modest number of queries (perhaps as many as ten thousand or more), one can learn
a classifier that closely mimics the black box ML service decisions.  Once
the black box has been reverse-engineered, the attacker need no longer subscribe to the
ML service.
Perhaps more importantly, such reverse-engineering enables TTE attacks by providing knowledge
of the classifier when it is not initially known. 
One weakness of \cite{Reiter} is that it neither considers very large (feature
space) domains nor very large neural networks -- orders
of magnitude more queries may be needed to reverse-engineer a DNN on a large-scale domain.  However, a more critical weakness of \cite{Reiter} 
which we will discuss in Section IV is that the queries in \cite{Reiter} are easily detected
because they are {\it random}, {\it i.e.} they do not use any knowledge of training samples for the domain. 

Recently, RE attacks
based on more realistic queries have been proposed \cite{Papernot3}.
First, the adversary collects a small set of representative labeled samples from the domain as an initial training set, $S_0$, and
uses this to train an initial surrogate classifier.
Then, there is stagewise data collection and retraining, over a sequence of stages. In each, the adversary augments the current training set by querying the classifier with the stage's newly generated samples, i.e.,
\begin{equation}
S_{k+1} =\{\underline{x}+\lambda\cdot{\rm sgn}(\nabla({\rm max}_cP^{(k)}_s[C=c|\underline{x}])):\underline{x}\in S_k\}\cup S_k
\label{sk}
\end{equation}
where $k$ is the current stage index and $P^{(k)}_s[C=c | \underline{x}]$ is the current surrogate class posterior model.
The surrogate classifier is then retrained using  the labeled $S_{k+1}$.  Based on (\ref{sk}), each successive stage crafts samples closer to the classifier's true
boundary, which is helpful for surrogate classifier learning (but which also makes these samples less class-representative and thus more detectable, as seen in Section V).
Once a sufficiently accurate surrogate classifier is learned, a TTE attack can be launched
using this classifier.


\subsection{Attacks on Privacy of Training Data}
Another emergent attack seeks not to alter classifier decisionmaking but rather to glean,
from the classifier, (assumed sensitive) information about the training set on which it was learned.
Relevant applications here include: discerning whether a particular person participated in a patient study that produced a disease classifier (or a diagnostic or prognostic decisionmaking aid) -- one may then infer he/she is likely to possess the disease; discerning whether a particular individual's data was used in training a system that grants secure access (to a building, data, financial records) only to company employees or vetted individuals.  One might instead seek to infer what such individuals {\it look like} (estimate an image of an employee's face \cite{Fredrickson15}).  There are various scenarios one can consider for this type of problem.  The most interesting (and realistic) one is wherein the attacker only has black-box (query) access to the classifier.  An interesting, representative paper that investigates data privacy attacks on classifiers under this scenario is \cite{Shokri17}.  This work makes some strong assumptions, but shows that when these assumptions hold one can make quite accurate inferences of whether or not an individual's sample was used in training (a membership inference attack), e.g., accuracies as surprisingly high as 80-90\% in inferring training-set membership on two classification domains.

The author's pose
the attacker's problem as learning a posterior model whose input is a data record (feature vector) and whose output is the probability that the data record was used in training the classifier under attack. There are three pivotal assumptions made: i) The attacker has query access to the victim classifier and, when queried, the classifier does not merely produce decisions - it gives decision ``confidence'' that could consist of the vector of class probabilities over all classes, just the ``top'' probabilities, or quantized values for these probabilities.
The attacker does not query the victim classifier repeatedly to reverse engineer its function -- this could be detected using the recent approach in \cite{ICASSP19}.  It simply queries using the data sample on which it is seeking to violate privacy, and elicits the victim classifier's decision and confidence on this sample; ii)  It is assumed that the attacker has access to a data set that is statistically similar to the training set used in building the victim 
classifier \cite{Shokri17}.  This assumption is plausible only in some applications.  In the patient study scenario, the attacker could have access to data records from hospital B, while the study yielding the victim classifier was produced by hospital A -- hospital B's population could be very similar to that of A.  However, it is less plausible the attacker would have a training
set statistically similar to a particular company's data set, used to build its secure authentification classifier -- unless {\it e.g.} members of this company also belong, in large numbers, to the same country club;
iii) Even though the victim classifier {\it is} assumed to be a black box with respect to the attacker, it is assumed that the 
victim classifier was trained using a particular on-line tool or ``ML pay-for-service'' system (e.g. provided by Google), that the attacker knows which tool/service was used, and he/she also has access to this tool.  In this way, even without knowing what the classifier type is (SVM, particular DNN),
the attacker can assume that, {\it given a similar data set, with the same feature vector format, from the same classification domain}, the tool/service is likely to produce a new classifier (a ``shadow'' classifier) that ``behaves in a similar way'' as the classifier under attack.  In particular, it should exhibit similar decision confidence patterns for samples used for training (high decision confidence), compared to samples not used for training.  
Using its own training and test data, the attacker uses the tool/service to build an array of such shadow models.
For each such model, he/she produces class decisions and the confidence scores on training samples, and separately on test samples.  Each such triple (class decision, confidence vector, train (Y/N)) is treated as a supervised instance of a {\it new} training set, used to learn a binary posterior model that infers whether or not a given sample was part of a shadow model's training set or not.  After the attacker's binary classifier is trained, it can be applied to the output of the classifier under attack, when queried by a given data sample, and yield the probability that the query sample was part of the attacked classifier's training set.

We have noted there are strong assumptions in this work, whose violation could substantially minimize the amount of membership ``leakage''
obtained.  First, if the victim classifier does not produce decision confidence, but merely a decision, this would defeat this attack.  Confidence on decisions is important, but one could {\it e.g.} grossly coarsen the victim classifier's output confidence to {``highly confident'', ``confident'', ``weakly confident'', ``uncertain''}.  Second, as the authors note, this attack is successful because trained classifiers tend to overfit to training examples, producing patterns of ``high confidence'' on samples used for training, and patterns of lower confidence for non-training samples.  The authors investigate defense strategies that seek to reduce classifier overfitting, or at least its signature in the victim classifier's posterior.  These mitigations include use of regularization and altering the victim classifier's posterior to increase its decision entropy.  While strong regularization can degrade the accuracy of the attacker's classifier, it may also compromise accuracy of the classifier under attack.  One can also likely defeat or weaken this attack by {\it not} using an accessible (and inferrable) service for training the victim classifier.  This attack may not transfer well if the victim and shadow classifier decision rules are quite different.

However, we have one insight that is not discussed in \cite{Shokri17}, and which leads to a simple procedure for defeating this attack.  {\it Simply suppose the following:  after training the targeted classifier, the training set that was used is retained by the platform/system that operates the classifier.  Now, when the classifier is queried by a sample, the system can first check if the sample is part of the training set or not.  If the sample is not part of the training set, the classifier can output its decision and confidence, as usual. However, if the query sample is in the training set (or even essentially indistinguishable from a training pattern, {\it i.e.} if the attacker added a small amount of noise to the query sample in order to be evasive), the system can infer that this is very likely a data privacy attack query}.  In this case, the classifier should still output the correct decision and confidence values that are at least MAP-consistent with that decision.  However, the system should randomize the confidence values to destroy any privacy revelation, and thus confound the attacker.  This simple defense should be highly effective at defeating \cite{Shokri17}.  Now, there are several potential arguments one can make against this defense.  First, perhaps the system should not retain the training set (since this makes it vulnerable to being stolen).  However, in many applications, there is great advantage to retaining the training set to allow future 
retraining (in light of more collected data, with the collection of additional discriminating features, etc.).  Second, as we noted, the attacker could add noise to his/her query so that it will not be recognized as a possible training instance.  However, there is a fundamental tradeoff between successful evasion of this defense and the attacker's success rate in revealing membership violations -- if insufficient noise is added, the system may infer that the likelihood this is an attack query is high.  If large enough noise is added to avoid such detection, the query sample may no longer elicit a high confidence ``signature'' from the classifier.  We suggest this tradeoff could be experimentally investigated in future.  We also note that the system can ignore differences from training samples in features that anyway are not crucial for class discrimination, as the attacker might obfuscate such irrelevant features to be evasive to detection while at the same time not altering the confidence pattern produced by the classifier -- in matching a query pattern against the training set, the detector should focus on the key, discriminating subset of the features.

\section{Review of Recent Work on AL Defenses}\label{sec:defense}
In this section we review recent work on defenses against TTE, DP, backdoor DP, and RE AL attacks.

\subsection{Robust Classification Approaches for TTEs}\label{sec:robust_class}


The robust classification approach aims to correctly classify any TTE examples, while maintaining high test accuracy in the absence of attacks. This is usually achieved by modifying the training (and/or testing) process -- preprocessing the data, using a robust training objective, or training with adversarial examples. Here we review these three robust classifier approaches.

\paragraph{Feature Obfuscation}
This approach alters features so as to hopefully destroy the attack pattern.
For example, \cite{Giles} considers DNNs for digit recognition and malware detection.  
They randomly {\it nullify} (zero) input features, both during training and testing/inference. This may ``eliminate'' perturbed features in a test-time pattern. However, for the malware domain, the features as defined in \cite{Giles} are {\it binary} $\in \{0,1\}$.  Thus, nullifying (zeroing) does not necessarily alter a feature's original value (if it is already zero). By first recoding the binary features to $\{-1,1\}$,  nullifying (zeroing) will always change the feature value.  This may improve the performance of \cite{Giles}. In \cite{Squeezing17} and \cite{LGH19}, 
quantization is used to destroy small perturbations.
However, if the perturbation size exceeds a quantization threshold, the perturbation could even be amplified, making the attack more effective. \cite{Li_ICCV} performs blurring of test patterns (images) in order to destroy a possible attacker's perturbations. This method will be experimentally evaluated in 
Section \ref{sec:robust_def}.
It was also shown in \cite{Boult} that some attacks are quite fragile and that even the process of image capture (cropping, etc.) may defeat the attack -- assuming the attack is on a physical object, not on an already digital image, image capture may give robustness against adversaries. Note that one common drawback of feature obfuscation is that a significant tradeoff may exist between accuracy in correctly classifying attacked examples and accuracy in the absence of attacks. Another problem, caused by involving discretizing modules, is the ``gradient masking'' effect \cite{Papernot3}. 
In principle such discretization, being non-differentiable, can defeat gradient-based TTE attacks.
However, gradient masking can be overcome by {\it transferred} attacks, obtained using a surrogate classifier whose gradients can be evaluated \cite{Papernot3, Carlini19}.

\paragraph{Incorporating Robustness into Training}
An early attempt here was made in \cite{Biggio}, which modifies the support vector machine (SVM) training objective to ensure the learned weight vector is not sparse, {\it i.e.} it makes use of all the features. Thus, if the attacker corrupts some features, other (unperturbed) features will still contribute to decisionmaking. However, \cite{Biggio} may fail if only a few features are strongly class-discriminating. For DNN classifiers, a distillation defense was proposed in \cite{Papernot2}, first training a DNN, and then using the class posteriors of each training pattern as ``soft'' labels to retrain the DNN. However, this defense was ineffective against the CW attack \cite{CW}. 
In {\it e.g.} \cite{Parseval17,Tsuzuku18}, the authors regularized the DNN training objective to control the local Lipschitz-continuity constant of each layer, so that perturbations added to input patterns will not significantly affect the output decisions through forward propagation. However, the local Lipschitz constant is specific to the type of layer, and may be difficult to 
precisely derive \cite{Pappas19}.
Several robust training approaches are designed to ensure decision consistency in a local region, {\it e.g.} an $L_2$ ball of radius $\epsilon$ around each training pattern as in  \cite{Provable,Tsuzuku18}.  
To this end,
\cite{Tsuzuku18} trains to achieve classification margin, {\it i.e.,} 
a lower bound on the difference between the score for the true class
and the largest score considering all other classes for all training samples.
Alternatively, such robustness 
may be 
achieved by first deriving an upper bound on adversarial loss in the region around the training examples ({\it i.e.} the supposed ``certified" region), 
and then minimizing the upper bound while training the classifier. 
For example, \cite{Certified} (focusing on single-layer classifiers) bounds the adversarial loss in the $L_{\infty}$ ball by bounding the gradient for all points in the ball. 
Compared with approaches that only consider the local gradient of each training pattern, \cite{Certified}  provides greater secured robustness since it suffers less from the gradient masking effect. 
But finding a tight upper bound on adversarial loss is a difficult optimization problem requiring heavy computation, especially for DNNs with a complicated structure \cite{BFT17, Weng18}; hence thus certified robustness is hard to achieve in practice. 
Also, such robust training does not guarantee a class-consistency 
region of prescribed size about nominal {\em test} samples 
and could reduce accuracy in absence of attacks \cite{Tsuzuku18,Certified}.
Especially for \cite{Tsuzuku18}, a large margin may result in an ``exploding gradient'', one of the categories of gradient masking effects, providing a false sense of robustness \cite{Athalye18}. 
Finally, neither \cite{Parseval17} nor \cite{Tsuzuku18} evaluated whether 
their classifiers are 
robust to TTE attacks based on transfer from a surrogate classifier, 
which is suggested in \cite{Carlini19} as a necessary evaluation step. 

\paragraph{Adversarial Robust Training}
First proposed in \cite{Szegedy_seminal}, an adversarially trained classifier is learned to minimize an {\it expected} training loss, where the expectation is over a distribution of bounded adversarial perturbations. This can also be seen as a type of DNN training with ``data augmentation'' ({\it e.g.}, to be robust to object position, scale, and orientation for digital images). Adversarially perturbed training patterns can be obtained by applying existing attacks, {\it e.g.} FGSM, CW, to clean training patterns using either the victim classifier \cite{KGB16} or a surrogate classifier \cite{Tramer18}. Compared with robust ``certified'' training approaches (with an $L_{\infty}$ ball), adversarial training uses an estimate of ``worst-case'' perturbations, which is less accurate but more efficient in computation. However, both approaches may fail
when the training and test data manifolds are different -- in such case both approaches are vulnerable to ``blind-spot'' attacks \cite{Zhang-blindspot}. In Section V, we will experimentally evaluate a state-of-the-art robust training approach
to assess its effectiveness against TTEs.

\subsection{Anomaly Detection (AD) of TTEs}\label{sec:AD-TTEs}

Over the past several years, there has been great interest in an alternative defense strategy against TTEs -- {\it anomaly detection} of  
these attacks, with a number of published works.
One motivation
given {\it e.g.} in \cite{Wagner17}, is that robust classification of attacked 
images is difficult, while detection is ``easier''.  
An argument in support of this is that if one designs a good robust classifier, then one gets a good detector essentially 
``for free'' -- one may make detections when the robust classifier's decision disagrees with a conventional (non-robust) classifier's decision.  
Likewise, considering methods like
\cite{Li_ICCV} which ``correct'' an image $X$ containing an attack, {\it e.g.} by blurring it, producing a new image $X'$, one can build a single classifier and make an attack detection when $\hat{C}(X) \neq \hat{C}(X')$.
However, such detection architectures, based on robust classification, are obviously not necessarily optimal.  In particular, 
considering the two classifier approach, if there is significant classifier error rate, then there are two significant sources for classifier non-consensus: i) a successful TTE attack on the conventional classifier and its correction by the robust classifier, and ii) misclassification of a non-attacked pattern by one (or both) of the classifiers.
In Section \ref{sec:expt}, we will show that direct design of an AD can yield much better detection accuracy than a detector built around robust classification.
Moreover, irrespective of whether detection is truly ``easier'', this argument (that detection is essentially a subset of robust classification) is not made in many of the robust classification defense papers -- it is simply assumed that
the only objective of interest is to defeat the attack by correctly classifying in the face of it.  Attack detection 
is not even considered in \cite{Biggio},\cite{Giles},\cite{Papernot2},\cite{Li_ICCV},\cite{IBM-Ireland}.
By contrast, in Section \ref{sec:dd}, we will give much stronger arguments for the intrinsic value in 
making detection inferences and will point out that, in some scenarios, when an attack is present, making robust classification inferences in fact has {\it no} utility.  
Moreover, we also argue that the attacker has a natural mechanism for {\it learning} a robust classifier (and then targeting it) -- querying -- whereas he/she
may have no ability to query so as to learn an AD that may be in play.
We defer more detailed arguments until Section \ref{sec:dd}. 

Irrespective of the motivation given, there has been substantial recent work on detection of TTEs.
Various approaches have been proposed, with a recent benchmark 
comparison study given
in \cite{Wagner17} evaluating a number of methods against the CW attack \cite{CW}, which
was demonstrated in \cite{CW},\cite{Wagner17} to be more difficult to detect than earlier attack methods such as
\cite{Goodfellow} and \cite{Papernot}.

\subsubsection{Supervised Detection of TTEs}

One strategy is to treat the detection problem as supervised, using labeled examples of ``known'' attacks.
The resulting binary classifier (attack vs. no attack) can then be experimentally evaluated both on
the known attacks and on unknown attacks.  Examples of such systems include \cite{Grosse} and
the supervised approach taken in \cite{AD}.  However, \cite{Grosse} failed
to detect the CW attack on the CIFAR-10 image domain
\cite{Wagner17}.  \cite{AD}
similarly proved unsuccessful in detecting CW on CIFAR-10 \cite{Wagner17}.  
\cite{Li_ICCV} also treated the problem as supervised, applying a multi-stage classifier, with each
stage working 
on features derived from a deep layer of the trained DNN classifier.  A detection is made unless
all the stages decide the image is attack-free. \cite{Wagner17} demonstrates this detector performs
very poorly on CW applied both to the MNIST and CIFAR-10 domains. \cite{Metzen}, which
feeds a DNN classifier's deep layers as features to a supervised detector, is 
more successful.  The best results reported for this supervised method in detecting CW
on CIFAR-10 were 81\% true positive rate (TPR) at a false positive rate (FPR) of 28\% \cite{Wagner17}\footnote{The FPR is the number of non-attack test patterns that are detected as attack divided by the number of non-attack test patterns.  The TPR is the number of attack test patterns detected as attack divided by the number of attack test patterns.}.
Later we will demonstrate an 
{\em unsupervised}
 AD which significantly exceeds these results.  Moreover, \cite{Metzen}
reported that training on some attacks did not {\it generalize} well to other attacks (treated as unknown).
This is an important limitation 
-- in general the defender may be proactive, without detailed knowledge of an attacker's
perturbation objective/strategy.

A related supervised approach is \cite{Safety-net}, which quantizes deep layer DNN features, feeding
the resulting codes as input to an SVM. This approach exploits gradient masking for attack evasion.  Again, being supervised, this approach may not 
generalize well to unknown attacks.

\subsubsection{Unsupervised AD for TTEs Without an Explicit Null Model}

Other approaches are unsupervised ADs, some of these based on explicit null hypothesis (no attack) 
statistical models for patterns, 
and others not forming an explicit null hypothesis.  
One crude approach is simply to
reject if the maximum
{\it a posteriori} class probability (produced by the classifier) is less than a given threshold.
Use of such ``confidence'' was shown to be effective for detection of a classifier's {\it misclassified} samples
in \cite{Hendrycks_misclass}, although effective detection of adversarial attacks may require more powerful
detectors.
Another non-parametric approach is based on principal component analysis (PCA) \cite{Hendrycks}.
However, \cite{Wagner17} found that, while successful on MNIST, this approach has essentially
no power to distinguish attacks from non-attacks on CIFAR-10.  

A more sophisticated, interesting non-parametric
detector is \cite{Magnet}. \cite{Magnet}
extracts nonlinear components via an auto-encoding neural network. 
It also uses somewhat unconventional decisionmaking, {\it combining} classification
and detection -- images whose (auto-encoding based) reconstruction error is large (far from the
image manifold) are detected as
attacks.  Images whose reconstruction error is small are sought to be correctly classified.
However, it may be desirable to make detections even when the
attack is subtle (see reasoning given in Section \ref{sec:dd}), with the attacked image lying close to the image manifold.  A primary concern with this 
detector is that it requires setting a number of hyperparameters (specifying the auto-encoding
network architecture, which in fact required different settings for MNIST and CIFAR-10, and softmax
``temperature'' variables).  Setting hyperparameters in an unsupervised AD setting is extremely difficult.
If labeled examples of an attack are available, one can set hyperparameters to maximize a labeled validation set
measure.  However, the attack is then no longer ``unknown'' and the detection method is actually
supervised.  \cite{Magnet} also incurs some degradation in classification accuracy in the 
absence of an attack -- from 90.6\% to 86.8\% on CIFAR-10 \cite{Magnet}.

\subsubsection{Unsupervised AD of TTEs based on an Explicit Null Hypothesis}
\begin{figure}[!htb]
\vspace{-0.15in}
\includegraphics[width=1\columnwidth]{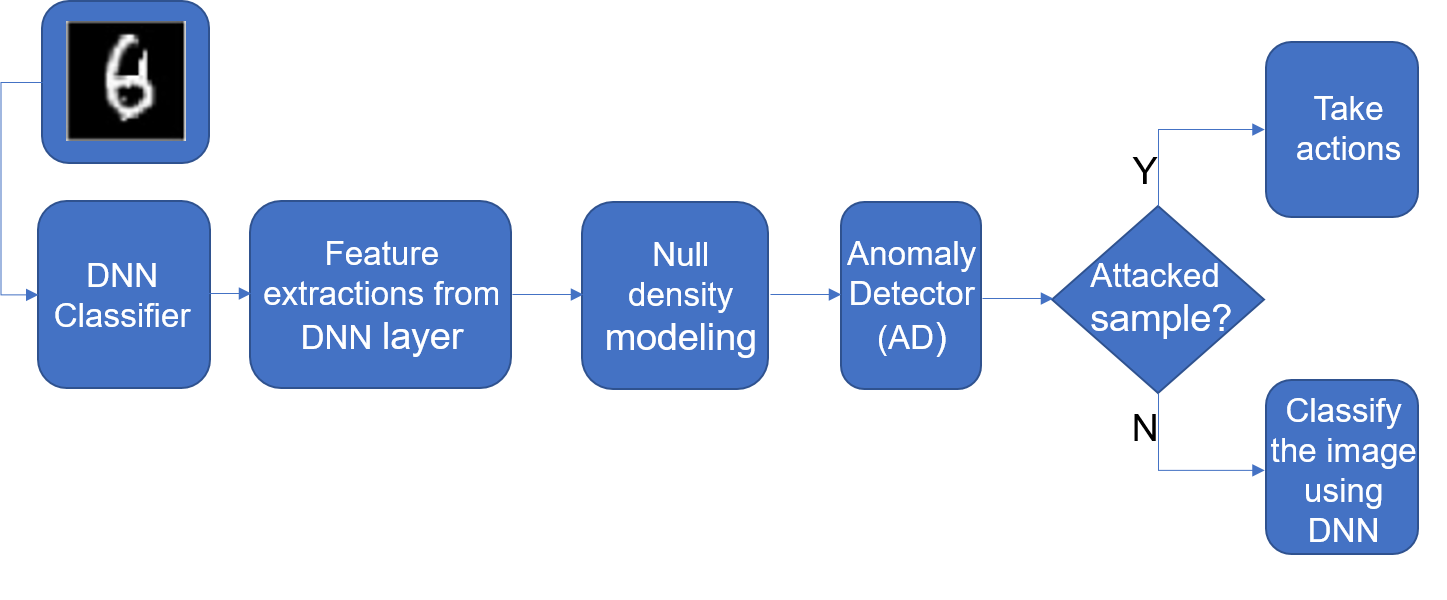}
\caption{Anomaly detection based on decision statistics derived from internal layers of a DNN.  Classification of the image is performed if no anomaly (attack) is detected.} \label{generic}
\vspace{-0.1in}
\end{figure}

A generic DNN-based detection plus classification system is shown in Figure \ref{generic},
with classification performed if no attack is detected.
One such detection approach with an explicit null hypothesis is \cite{Openmax}.  This approach computes the distance
between a DNN layer's class-conditional mean feature vector and the test image's feature vector
and then evaluates this, under the null hypothesis of no attack, using a Weibull distribution.
A few limitations of \cite{Openmax} are that: i) it does not model the {\it joint} density of a deep
layer -- such a model would exploit more information than just the (scalar) distance; ii) \cite{Openmax} is not truly
unsupervised -- several hyperparameters are set 
by maximizing a validation measure that requires 
labeled examples of the 
attack.  The experiments of Section \ref{sec:expt} will show that a purely unsupervised AD \cite{MLSP18-ADA}  outperforms \cite{Openmax}
even though we (optimistically) allow \cite{Openmax}'s detector to use more than one hundred labeled examples of the 
attack to be detected.

More recently, \cite{AD} did propose a method based on a null hypothesis joint density
model of a DNN's deep layer feature vector.  To our knowledge, theirs is the first such approach,
upon which the detection methodology in \cite{MLSP18-ADA} builds.  \cite{AD} used a kernel density estimator to
model the penultimate layer of a DNN.  However, ultimately, they put forward a {\it supervised}
method, learning to discriminate `attack' from `no attack', leveraging their density model to create the classifier's input features.  They ultimately settled on a supervised method because their
unsupervised detector did not achieve very good results.  In \cite{Wagner17}, it was found that
the unsupervised AD in \cite{AD} 
grossly fails on CW attacks on CIFAR-10 -- 80\% of the
time, attacked images had even {\it higher} likelihood under the null model in \cite{AD} than the original (unattacked)
images.
The unsupervised `anomaly detection of attacks' (ADA) \cite{MLSP18-ADA} method 
that
builds on \cite{AD} is a much more effective detector, one which currently achieves state-of-the-art
results.  We next give a brief description of this approach.

\begin{figure}[h]
\centering
\includegraphics[width=0.5\columnwidth]{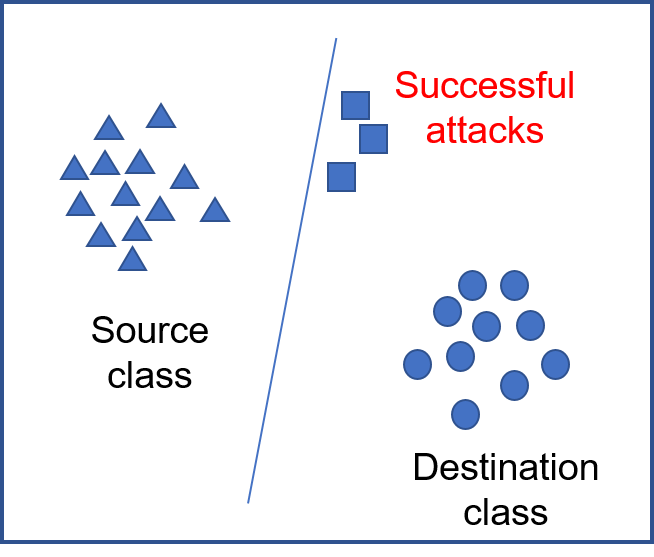}
\caption{Illustrative attack examples in the plane with a linear decision boundary -- such examples may have relatively high likelihood under the source class and low likelihood under the attack's destination
class.}
\label{fig:source-dest}
\end{figure}

\noindent
{\bf Summary of the ADA Method:}
Consider a successful, targeted TTE attack example -- one which was obtained by starting from a ``clean''
example $\underline{x}$ from some 
source class $c_s$ (unknown to the defender) and then perturbing it until the DNN's decision on this perturbed example (now $\underline{x'}$)
is no longer $c_s$, but is now $c_d \neq c_s$ (the ``destination'' class).  The premise behind \cite{AD} is that feeding an {\it attacked} version of $\underline{x}$, not $\underline{x}$ itself, into the DNN and extracting, as the derived feature vector $\underline{z}$, the vector of outputs from some internal layer
of the DNN ({\it e.g.}, the penultimate layer, right before the decision layer)
will result in a feature vector $\underline{z}$ that has atypically low likelihood under a learned {\it null} density model for 
$\underline{z}$ ({\it i.e.}, the estimated distribution for $\underline{z}$ when no attack
is present), conditioned on the DNN-predicted class $c_d = c^{\ast}$.
Denote these null densities, conditioned on each class $c$, by $f_{\underline{Z}|c}(\underline{z}|c)$.
If the perturbation of $\underline{x}$ is not very large (so that the attack is human-imperceptible and/or detection-evasive),
one might {\it also} expect that $\underline{z}$ will exhibit
{\it too much typicality} (too high a likelihood) under some class other than $c^{\ast}$, {\it i.e.}
under the {\it source} category, $c_s$.  This additional information is exploited in \cite{MLSP18-ADA}. It does not matter that the source category is unknown to the defender.  He/She can
simply determine the best estimate of this category, e.g. as: 
$${\hat c}_s = \arg\max_{c \in \{1,\ldots,K\} - c^{\ast}}
f_{\underline{Z}|c}(\underline{z}|c).$$
Accordingly, it was hypothesized that attack patterns will be {\it both} ``too atypical'' under $c^{\ast}$
and ``too typical'' under ${\hat c}_s$.  
This is illustrated in Figure \ref{fig:source-dest}.

While this may seem to entail an unwieldy detection strategy that
requires use of two detection thresholds, a single, theoretically-grounded
decision statistic that captures both requirements (assessing ``atypicality'' with 
respect to $c^\ast$ and ``typicality'' with respect to ${\hat c}_s$) was proposed.
Specifically, define a two-class posterior evaluated with respect to the (density-based) null model:
$$P \equiv [P_{c^{\ast}},P_{\hat{c}_s}] = [p_0 f_{\underline{Z}|c^{\ast}}(\underline{z}|c^{\ast}),p_0 f_{\underline{Z}|{\hat c}_s}(\underline{z}|{\hat c}_s)],$$
where $p_0$ gives the proper normalization: 
$$p_0 = (f_{\underline{Z}|c^{\ast}}(\underline{z}|c^{\ast}) + f_{\underline{Z}|{\hat c}_s}(\underline{z}|{\hat c}_s))^{-1}.$$
Likewise, define the corresponding two-class posterior evaluated via the DNN:
\begin{eqnarray*}
Q & \equiv & [Q_{c^{\ast}},Q_{\hat{c}_s}]\\
&  = &  [q_0 P_{\rm DNN}[C= c^{\ast} | \underline{x}], q_0 P_{\rm DNN}[C = {\hat c}_s | \underline{x}]],
\end{eqnarray*}
where 
$$q_0 = (P_{\rm DNN}[C= c^{\ast} | \underline{x}] + P_{\rm DNN}[C= {\hat c}_s | \underline{x}])^{-1}.$$
Both deviations (``too atypical'' and ``too typical'') are captured in the Kullback-Leibler divergence 
decision statistic: 
$$D_{\rm KL}(P || Q) = \sum\limits_{c \in \{c^{\ast},\hat{c}_s\}} P_c\log(\frac{P_c}{Q_c}),$$
 {\it i.e.} a detection is declared when this statistic exceeds a
preset threshold, chosen to meet a targeted FPR. 

We emphasize that the hypothesis that an attacked pattern will exhibit both ``too high atypicality'' (with respect to $c^{\ast}$) and
``too high typicality'' (with respect to ${\hat c}_s$) is not based on any strong assumptions about the
attacker.
It merely assumes the perturbation is constrained to be small to be human-imperceptible and/or machine-detection evasive. 
This constraint, in conjunction with the attacked example's misclassification by the classifier, 
may necessitate that the test pattern will exhibit unusually high
likelihood for a category ($c_s$) other than that decided by the DNN.  
It may also necessitate that the test
pattern will exhibit unusually {\it low} likelihood under $c^{\ast}$ (The perturbation is not
large enough to make the perturbed image a typical pattern of $c^{\ast}$.).

The baseline ADA method was further improved by:
i)
modeling the joint density of a deep layer using highly suitable mixture density-based null hypothesis
models (in particular, through log-normal mixture density modeling to match to the {\it non-negative support} for RELU layers); ii)
exploiting {\it multiple} DNN layers by taking the {\it maximum} of the ADA statistic over each of the layers being
evaluated;
iii) accounting for uncertainty about the source class by computing an expected ADA statistic, based on the estimated probabilities for each of the classes being the source class;
iv)
exploiting the class confusion matrix, comprehensive {\it low-order} density modelling \cite{Qiu16}, and DNN weight information in constructing
an {\it ultimate}, best-performing, ADA statistic \cite{MLSP18-ADA}
which demonstrated state of the art results against published attacks,
 including CW \cite{CW}.  \cite{MLSP18-ADA} directly
compared against several benchmark detection methods and also against the
results reported in \cite{Wagner17}.
Also evaluated were some performance measures of great 
interest that are sometimes not assessed
in prior works, including multiple performance measures as a function of attack strength and 
classification accuracy (in the absence of attacks)
versus the false positive rate of attack detection.
Section \ref{sec:expt} gives an evaluation
of  ADA against other detectors for several classification
domains.
\subsection{Defenses Against Classifier-Degrading DP Attacks}
The dominant objective in defending against {\it embedded} DP attacks on classifiers is some form of robust
learning, aiming to preserve (generalization) accuracy in the presence of training set poisoning.
Two common such training strategies are as follows:
i) performing some type of training set {\it sanitization}, followed by standard classifier learning.  Sanitization could consist of detection and removal of outlier samples.  Alternatively, it could involve 
feature selection or compaction ({\it e.g.}, principal component analysis, followed by removal of low energy
coefficients), intended to neuter data poisoning by removing minor signal components/features, presumed to be primarily used by the attacker;
ii) robustly modifying the classifier training objective function, accounting both for outliers as well as
{\it inliers} that are adversarially mislabeled.  
As an example of ii), \cite{Xiao15} accounts for
both random as well as adversarial mislabeling for kernel-based SVMs by replacing the dual SVM objective function, which depends on the training labels, with an {\it expected} dual function, based on the probability of mislabeling.  While this provides robustness benefits, it is not surprising that similar benefits are also obtained
in \cite{Xiao15} by reducing the SVM (slackness) penalty on margin violations.  This allows more data points to contribute to the estimation of the SVM's weight vector as support vectors, which provides inherent robustness.  Another issue is that the solution in \cite{Xiao15} relies on knowledge of the mislabeling probability, {\it i.e.} a (generally unknown) hyperparameter.  While one could choose this parameter in a conservative, worst-case fashion (assuming a high mislabeling probability), this will compromise accuracy in the 
{\it absence} of a DP attack. 

\cite{Steinhardt17} considers the first strategy, performing (nearest neighbor based)
outlier detection followed by SVM optimization.  Under certain assumptions (in particular, that removal of
{\it non-attack} outliers will not have a strong impact on estimation of the classifier), they are able to
obtain a (domain-dependent) upper bound on the worst case loss in accuracy from a DP attack.  Subsequently,
in \cite{Koh18}, the same authors considered reactive attacks against the robust classifier defense considered in their
earlier work \cite{Steinhardt17}.  This latter work reports that DP attacks with knowledge
of the sanitization defense can {\it evade} it and significantly degrade accuracy.
In particular, the attack may relatively densely concentrate poisoned samples in close proximity, so that they do not resemble outliers.

One crucial assumption in \cite{Koh18} is that the attacker has full knowledge not only of the training set {\it but also the test set}, {\it i.e.} the set of samples on which the authors evaluate generalization loss.
The attacker's objective function, to be maximized with respect to the chosen poisoned training samples,
is the classification loss {\it measured} on this test set.  Thus, the same test set is used by the 
attacker in crafting his/her attack {\it and} by the authors to evaluate the degradation in accuracy induced by this attack.  This appears to be highly biased -- an independent test set, not used by the attacker,
is needed to evaluate the true loss in accuracy associated with this attack. Note that merely knowing the training set, the attacker already possesses an advantage over the defender (since he/she knows the {\it clean}
training set, without the poisoned samples).
We accordingly suggest that the approach in \cite{Koh18} could be investigated with the adversary's attack objective function
based on a {\it surrogate} for the test set -- {\it e.g.}, the clean training set -- rather than the actual (test) set of samples used for assessing the accuracy degradation induced by the attack.
Introducing poisoned samples that induce large training set loss should {\it also} induce an increase in (unseen)
generalization error (estimated based on a held-out test set).
We also note that the attacker could also/instead poison the data set used by the defender to {\it estimate} the classifier's accuracy ({\it i.e.}, a finite validation or test set),
seeking to make this estimate inaccurate. 
This is yet another DP attacker objective.

There is also prior work on DP defense specific to email spam filtering and to the (less challenging)
{\it on-line} DP scenario, rather than the embedded DP scenario.  
Naive Bayes spam filters are highly susceptible to DP 
attacks.  Here, known spam sources/blacklisted IP addresses exploit the fact that
their received emails will be treated as (ground truth) labeled spam examples, and used for classifier training (or re-training).  The attacking source thus generates emails that will skew the
spam model, potentially resulting in great degradation in classifier accuracy. One approach for defending
against this is proposed in 
\cite{Roni}, where it is assumed that: i) one starts with a clean spam-ham training set; ii) there is a clean spam-ham validation set;
and iii) poisoned spam is added to the training set {\it subsequently}, {\it e.g.} in an on-line
setting.  Under these assumptions, it is not difficult
to build a detector based on degradation in the (on-line) learned spam classifier's accuracy on the validation set -- if adding a labeled spam example (or a small batch of such samples) to the training set results
in degradation in validation set accuracy, one can reject these samples as poisoned.  However,
{\it more generally}, if the DP attack samples are embedded as part of the {\it original} training set and if there are no time stamps, data provenance,
or other means for narrowing to a particular subset of the training set as possibly anomalous,
the problem is much more challenging.
For the approach in \cite{Roni}, under this embedded attack scenario, there would be a combinatoric explosion in the set of candidate poisoned subsets of the training set to assess, which makes \cite{Roni} quite impractical for addressing embedded attacks. 

Defense against 
DP attacks applied to active learning has also
been considered \cite{MLSP17}.
Active learning systems efficiently {\it grow} the number of labeled training samples by selecting, for
labeling by an ``oracle'' ({\it e.g.}, human expert), the samples from an unlabeled batch whose labelings are expected to improve
the classifier's accuracy the most.  The classifier is retrained with each new labeled sample (or new mini-batch of labeled samples, if samples are labeled more than one at a time).
\cite{Tygar14} has shown that active learning of classifiers can be compromised by adding poisoned samples to
the unlabeled batch on the current classifier's decision boundary {\it if} the oracle provides somewhat noisy (adversarial) labels for these samples\footnote{The oracle is usually assumed to provide accurate, ground-truth labels for samples actively selected for labeling.  However, it is possible that the oracle/human analyst is unreliable or is an insider.  In such settings, the assumption made in \cite{Tygar14} is reasonable.}.
Active learning methods typically select for oracle (human analyst) labeling the sample with highest decision
uncertainty ({\it i.e.}, a sample 
close to the current boundary).  This plays right into the adversary's hands, as the poisoned samples will often be selected for labeling.
\cite{MLSP17} proposed a defense against this type of attack which exploits randomness to good effect.  
Note that, anyway, there
are multiple active sample selection criteria of operational interest: i)  highest
class-entropy (decision uncertainty); ii) most outlying sample (potentially an instance of an {\it unknown} class, {\it i.e.} with no labeled examples \cite{Qiu16}); iii) random selection; iv) selecting the sample most likely to belong to a category of interest, {\it e.g.} the most actionable category, with confirmation by the oracle.  Thus, \cite{MLSP17} proposed a {\it mixed} sample selection strategy, randomly selecting from multiple criteria at each oracle labeling step.  This approach reduces the frequency with which an adversarial sample is selected for labeling and thus mitigates accuracy degradation associated with data poisoning.  It also ``services'' multiple active learning objectives.  As seen later (and also earlier, in \cite{Giles}), randomization plays an important role more generally in some defensive schemes against adversarial attacks. 

Finally, a related problem is where the focus of adversaries 
is on compromising the ground-truth labeling mechanism.  
In particular, a problem of great interest is adversarial {\it crowdsourcing}.
In crowdsourcing systems for obtaining ground-truth labels, there are a set of ({\it e.g.} Amazon Mturk) workers who each make decisions on a given example (news article, image, etc.), with the decisions across the worker ensemble aggregated to produce estimated
``ground truth''.  Such labeling could be used to create a supervised training set for learning a classifier.  Alternatively,
it could also be used as an {\it alternative} to an automated classifier for making decisions (and taking actions) on a given
set of (operational) patterns.
Adversarial crowdsourcing means that some of the workers may be giving unreliable answers, either because they are lazy/unskilled or because, in an intentional way, they want to {\it alter} the aggregate decision that is reached.
One can accordingly pose robust aggregation schemes that attempt to defeat the adversaries and perhaps even identify them
\cite{Karger11}, \cite{Kurve15}.  In \cite{Kurve15}, a stochastic model for workers is developed and an
associated EM algorithm learns, in a transductive setting\footnote{In transductive learning, the learner does not build a classifier model from training data with the future goal of applying this classifier to make decisions on operational (test) data.  Instead, the learner's objective function is over the (batch of) {\it test} samples, with the 
learner's optimization directly yielding the (hard or soft (class posterior)) decisions on these samples.}, from a batch of workers' answers, aggregate posterior
probabilities for the possible answers for each of the given tasks in the batch.  Moreover, the algorithm learns parameter values that characterize
each individual worker's skill level, the relative difficulty of each of the tasks, {\it and} each worker's intention (honest or adversarial).  It was demonstrated that this approach is quite robust to adversarial workers amongst the crowd (it can defeat the adversary's goal of altering the aggregate decision) and it can also reveal adversarial workers (who could then be blocked from participating).  Moreover, the method was found to be effective both in the semisupervised setting, where some ``probe'' tasks have ground truth answers that help to reveal adversaries, {\it as well as} in a purely
unsupervised setting.

\subsection{Defenses Against Backdoor DP Attacks}
While backdoor attacks on classifiers were only proposed fairly recently \cite{Song}, there is already nascent 
activity aiming to detect and mitigate them \cite{KLiu18},\cite{BChen18},\cite{Madry-NIPS18},\cite{MLSP19-backdoor},\cite{backdoor-perceptible,TrojAI}.  
There are three scenarios here: i) detecting the attack in the training set, before/during classifier training; ii) detecting backdoors ``in-flight", {\it i.e.} during the classifier's use/operation, when it is presented with a test example (that may contain a backdoor) to classify; iii) intriguingly, detecting backdoors {\it post-training}, {\it without access to the (possibly poisoned) training set}.
\subsubsection{Detection Using the Training Set and Trained DNN}
\cite{KLiu18}
considers a scenario where a user outsources classifier training to a third party, who may be adversarial.
The user provides a (presumably) clean training set to the third party {\it as well as} a specification
of the hyperparameters of the DNN (the number of layers and the sizes of these layers).  The adversarial third party adds backdoor training patterns, with target class labels, to the training set.  The adversary will seek the classifier to learn high accuracy on ``normal'' patterns, but while classifying to the targeted class when the backdoor pattern is present.  
Crucial to \cite{KLiu18} is that the user/defender also possesses a clean validation set, not available to the attacker, which can be exploited for defense purposes.  The premise behind this work is that  
backdoor patterns will activate neurons that are not triggered by clean patterns.  Thus, the defender can prune
neurons in increasing order of their average activations over the (clean) validation set, doing so up until the point where there is an unacceptable loss in classification accuracy on the validation set.  This is likely to remove the neurons which trigger on backdoor patterns.  The authors' assumption that the user specifies the DNN hyperparameters to the third party ensures that the network is made large enough to allow subsequent pruning without loss in accuracy -- if the third party had control of the network architecture, he/she could make it compact enough that
{\it any} pruning would necessarily result in loss in classification accuracy.  The scenario of this work is somewhat artificial -- the choice of the network (its size) is a fundamental aspect of model learning.  Pre-specification of this by the user (who is {\it not} the learner in this case) may not ensure good accuracy is achievable (the user could underspecify the model size).  Moreover, while in this scenario the user is supplying the data and could ``hold back'' a clean validation set, the {\it embedded} data scenario is of great interest not only for general DP attacks but also for backdoor attacks. In this challenging scenario, any held out data may {\it also} contain backdoor attack patterns. Finally, there is nothing about standard DNN training that should create a propensity for neuron ``dichotomization'', {\it i.e.} the existence of neurons whose sole use is to implement the backdoor mapping.

\cite{BChen18} does address the embedded scenario, where there is no clean validation set available.
The authors consider an internal layer ({\it e.g.} the penultimate layer) of a DNN, extracting this as a feature vector for an input training pattern.  They first perform dimensionality reduction on this vector (using 
independent component analysis) and then perform K-means clustering on the resulting training set feature vector patterns all labeled to the same category, seeking to separate each class's samples into two clusters.  They then propose three different ``inference''
strategies to determine whether one of the two clusters is a ``backdoor'' cluster. The authors do suggest their approach is successful on MNIST even when {\it clean} classes are multi-modal (requiring multiple clusters to represent them) -- this was experimentally validated by pooling several digits and treating them as one class. It is unclear whether these results would hold on more complex data sets.  Some limitations of \cite{BChen18} are that: i) it is assumed backdoor patterns manifest in principal components -- if a pattern manifests in minor components, the front-end dimensionality reduction may destroy the backdoor pattern prior to the clustering step; ii) it relies on the effectiveness of K-means clustering, which is susceptible to finding poor local optima; iii) it may not be so effective in general when classes are multi-modal (in such a case, a two-clustering may not isolate the backdoor attack in one cluster). 

\cite{Madry-NIPS18} does not rely on clustering per se to detect backdoors.  Rather, the authors show that if a backdoor 
pattern appreciably alters the mean feature vector pattern of a class (with feature vectors again obtained from the penultimate layer of the DNN, when the image is applied as input to the DNN), then with high probability clean and backdoor patterns labeled to this class can be separated by projecting onto the principal eigenvector of the feature vector's covariance matrix (This amounts to clustering in one dimension).  Again, classifier retraining follows removal of backdoor training samples.
The authors demonstrate promising results for this approach applied to the CIFAR-10 data set.  We will further investigate this method experimentally in section V. 

One issue with
\cite{Madry-NIPS18} is that the detection threshold is based on knowledge of a hyperparameter that essentially bounds the number of poisoned training patterns.  In practice, this number will be unknown. While the authors select this guided by test set accuracy, backdoor patterns could also be embedded in the test set (to confound use of the test set for setting this hyperparameter).
Thus, one cannot in general use (good) test set accuracy as a guide for setting this
hyperparameter. Second, while the authors do demonstrate both that their approach identifies most of the backdoor patterns and, in removing them, retains high classification accuracy, they do not report how many clean patterns are falsely rejected by their method.  For CIFAR-10 there is a sufficient amount of training examples (5000 per class) that accuracy should be unaffected even if many clean patterns are falsely removed.  However, this could be more problematic if fewer training samples are available. 
Moreover, one can envision backdoor pattern mechanisms that may not induce large changes in mean pattern vectors.  In such a case, one might need to scrutinize minor components, as well as the principal components.
\subsubsection{In-Flight Detection}
This is a nascent detection area.
Here, we mention \cite{STRIP}. The proposed method mildly assumes
availability of a small batch of clean patterns (from multiple classes) from the domain, $\{\underline{x}_i\}$.  For a given test pattern
$\underline{x}$, they form $\underline{x} + \underline{x}_i$ and classify it by the trained DNN.  
They then measure the class decision entropy over this small batch.  If a backdoor is present, entropy should be low
(the classifier decision is mainly made based on presence of the backdoor pattern).  If $\underline{x}$ is clean, the decision entropy is expected to be higher.  \cite{STRIP} demonstrates good detection results on both MNIST and CIFAR-10 when the attack strength is high (a visible black pattern superimposed on the image).  It is unclear how
effective this method will be as the attack strength is reduced.
\subsubsection{Post-Training Detection Without Access to the Poisoned Training Set}
This is a highly intriguing recent problem practically motivated because {\it e.g.} a trained classifier may be the basis of a phone app that will be shared with many users or may be part of legacy
national or industrial infrastructure -- in many such cases, the recipient (user) of the classifier may not have access to the
data on which it was trained. Detecting backdoors post-training, without access to the poisoned
training set, may thus reveal and thwart highly
consequential attacks. 
However, without the training set, the only potential hint of a possibly embedded backdoor pattern is {\it latent} in the learned DNN's weights.  Recently, highly promising work has addressed this problem under two scenarios that collectively cover the evasive backdoor cases of interest:
i) where the backdoor pattern was made evasive by being imperceptible (either modifying a few pixels or a global, image-wide watermark pattern); ii) where the backdoor pattern is {\it perceptible but innocuous}, {\it e.g.} a ball placed in front of a dog, whose presence causes the
classifier to misclassify a dog as a cat, or a pair of glasses on a face, which alters the class decision.  In \cite{TrojAI}, the first scenario was addressed by a 
purely unsupervised AD defense that: 1) detects whether the trained DNN has been backdoor-attacked; 2) infers the involved source and target classes; 3) it was even demonstrated that it is possible to accurately estimate the backdoor pattern. 
The AD approach involves {\it learning}, using a data set of labeled, unpoisoned examples from the domain, via suitable cost function minimization, the minimum size perturbation ({\it putative} backdoor) required to induce the classifier to misclassify (most) examples from class $s$ to class $t$, for all $(s,t)$ pairs. {\it For imperceptible backdoors, the hypothesis is that non-attacked pairs require large perturbations, while attacked pairs require much smaller ones}.  Thus, an order statistic p-value based hypothesis testing approach is invoked to assess how atypical is the perturbation size/norm for the class pair with the {\it smallest} perturbation 
achieving a high level of group misclassification, compared to that of all other class pairs.  The underlying hypothesis, and the detection strength of the resulting approach, are convincingly borne out by the results in \cite{TrojAI}.  Some results for this method are provided here in Section \ref{sec:expt}.  For innocuous, perceptible backdoors, the 
underlying hypothesis is that {\it if the spatial support of the estimated backdoor patttern for
	a class pair $(s,t)$ is restricted in size}, high group misclassification is {\it only} achievable for a class pair $(s,t)$ truly involved in a backdoor mapping.  Promising experimental results for this
approach are provided in \cite{backdoor-perceptible}.
\subsection{Defenses Against RE Attacks}
There is a paucity of prior art on defending against RE attacks, {\it i.e.} this is a nascent 
research area.  One interesting idea, developed for SVMs,
is \cite{cikm14}, which makes the {\it classifier} evasive
(a moving target), rather than the adversary.  That is, the authors
propose a {\it randomized} classifier, adding multivariate Gaussian noise to the classifier's weight vector,
with the covariance matrix chosen to ensure both high classifier accuracy and as much ``spread'' in weight vector realizations as possible.  This is nicely formulated as a convex optimization problem.  One limitation of 
\cite{cikm14} is that this approach may not naturally extend to DNNs.  A more serious concern is that the authors
do not show that randomization prevents an RE attacker from learning a good classifier for the domain -- they 
only show that RE-learned weight vectors deviate significantly from the mean SVM weight vector.  Moreover,
the authors evaluate the success rate of TTE attacks following the RE classifier learning phase.  While TTE success rates are reduced by classifier randomization compared with a fixed classifier, the TTE success rate is still quite high -- {\it e.g.} dropping from 80\% to 65\% on some data sets.  The problem here is that because 
the randomization retains high classification accuracy, RE querying, even of {\it random} classifier realizations,
acquires generally accurate labels for its queries.

An alternative approach for RE defense, proposed recently in \cite{ICASSP19}, is to {\it detect} RE querying,
much in the way that one can detect TTE querying.  That is, because
the attacker does not know that much about the classifier (and may have limited legitimate examples from the
domain), he/she will have to be ``exploratory'' with queries.  
However, one can then hypothesize that
query patterns may be atypical of legitimate examples from the domain, much as are TTE attack examples.  In
fact, \cite{ICASSP19} directly leverages the ADA detector, designed to detect TTE attacks, to detect RE querying.
In \cite{ICASSP19}, it is shown that this approach achieves high ROC AUC (0.97 or higher) in detecting 
the RE querying proposed in \cite{Papernot3} for MNIST, and at a ``stage'' of querying {\it before} the attacker has learned
an accurate classifier, suitable for mounting a TTE attack.  In Section 
\ref{sec:expt}, we will extend these results to
also demonstrate success in RE detection on the CIFAR-10 domain.

\section{Digging Deeper: Novel Insights and Broad Perspectives, Particularly About TTE Attacks} \label{sec:dd}

\subsection{Defenses Against TTEs: Robust Classification versus Attack Detection} \label{sec:robust_def}

As reviewed earlier in Section \ref{sec:robust_class}, 
some TTE defenses simply seek to correctly classify TTE attack examples. 
Robust classification
is an important goal, especially in the presence of ``natural'' adversaries --
additive noise, transformations (image rotation, translation, and
scaling, variable lighting), etc.  A common approach in training DNNs is data augmentation, with pattern variants
added to the training set so that the trained classifier is as robust as possible.
However, the works \cite{Biggio},\cite{Giles},\cite{Papernot2},\cite{Li_ICCV},\cite{IBM-Ireland}
do not focus on (even consider) achieving robustness to natural adversaries, but rather solely to TTEs. 

\begin{figure}[h]
\centering
\includegraphics[width=\columnwidth]{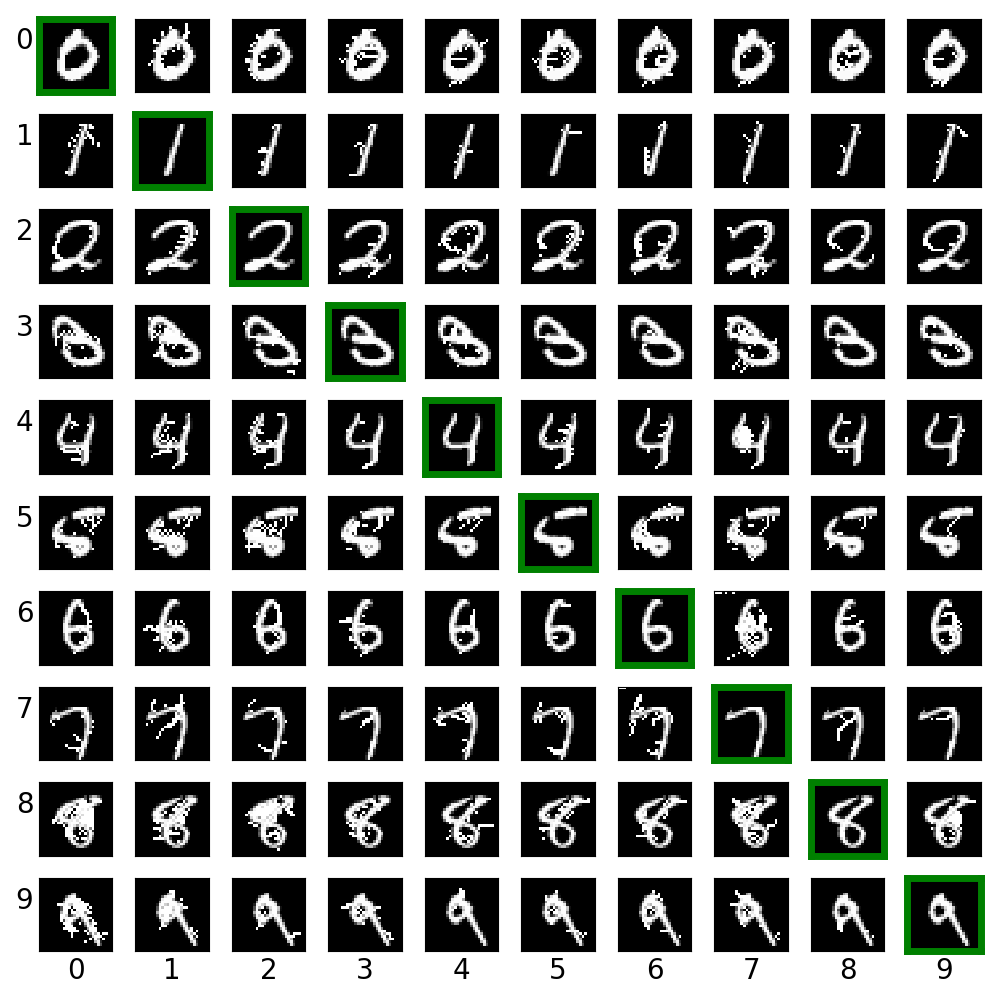}
\caption{JSMA \cite{Papernot} adversarial image matrix, with the row the true class and with the column the decided class.  The diagonal consists of unattacked digits. 
}\label{fig:JSMA-matrix} 
\end{figure}

One limitation of robust classification to combat TTEs 
concerns semantics of inferences.  Consider digit recognition.
Evasion-resistance means a perturbed version of `5' is still classified as a `5'.
This {\it may} make sense if the perturbed digit
is still objectively recognizable ({\it e.g.}, by a human) as a `5'.
(Even in this case, the decision may have no legitimate
utility.  This will be explained in the sequel.)
However, the perturbed example may not be unambiguously recognizable as a `5' --
consider the 
perturbed digit examples based on the attack from \cite{Papernot} shown in 
Figure \ref{fig:JSMA-matrix}, with significant salt
and pepper noise and other artifacts.  
Some attacked examples from classes `5', `3', and `8' are noticeably ambiguous.
This suggests, at a minimum, that there should be a ``don't know'' category for domains susceptible to TTE attacks.
However, irrespective of 
assigning the pattern to the correct class (which might in fact be ``don't know''),
it may be
crucial {\it operationally} to infer that the classifier is being subjected to an evasion attack.  Robust classification does not make such inference.
Once an attack is detected, measures to defeat the attack may be taken -- e.g., blocking
the access of the attacker to the classifier through the
use of multifactor (challenge-response) authentication.  
Moreover,
actions typically made based on the classifier's decisions may either be 
preempted or conservatively modified.  
For example, for an autonomous vehicle, once an attack on its
image recognition system is detected,
the vehicle may take the following action sequence: 1) slow down, move to the side of the road, and stop;
2) await further instructions. In a medical diagnostic setting, with an automated classifier
used {\it e.g.} for pre-screening numerous scans, a radiologist should
obviously not try to make a diagnosis based upon an attacked image 
 -- if the attack is detected
the patient can be re-scanned.  In \cite{Giles},\cite{Biggio},\cite{Papernot2} and
other papers, it is presumed that correctly classifying attacked test patterns is the right objective, without
considering attack detection as a separate, important inference objective. 

\begin{figure}[h]\centering
\centering
\includegraphics[width=\columnwidth]{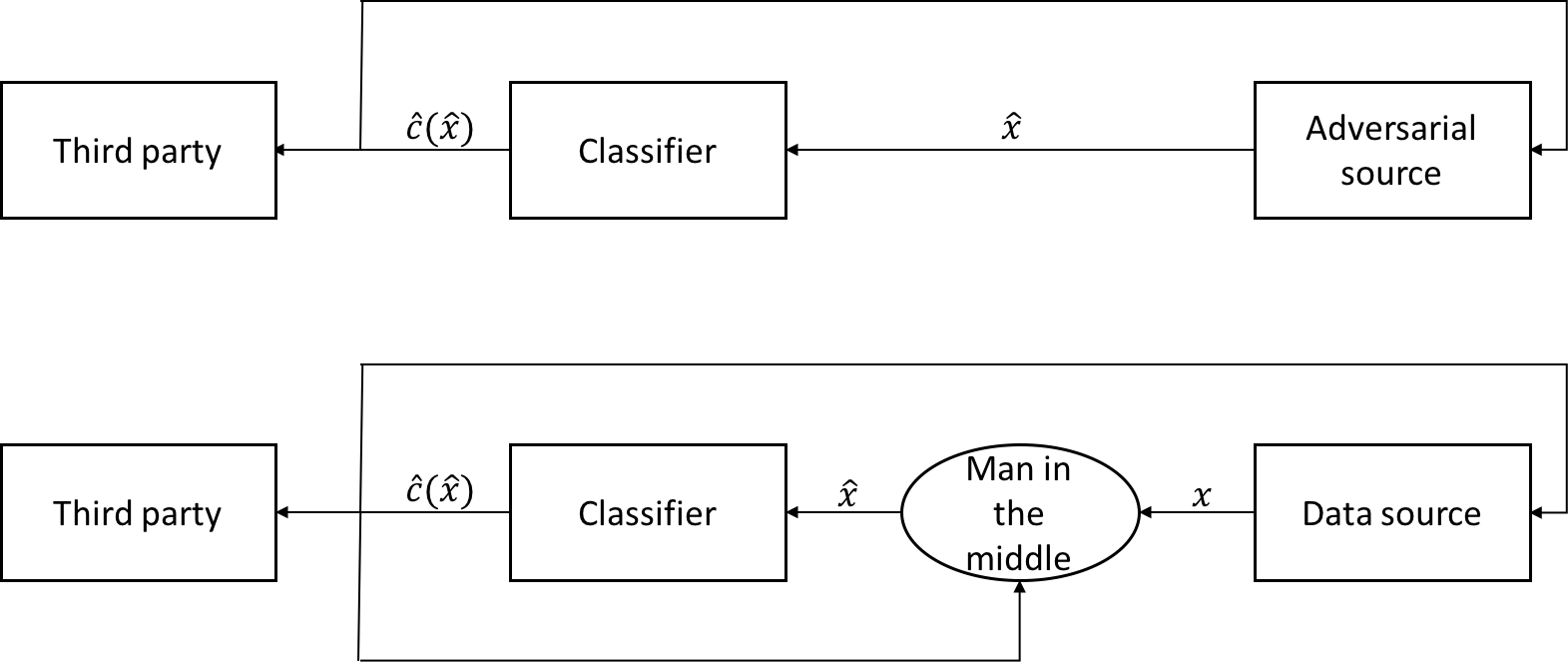}
\caption{The two possible test-time evasion attack mechanisms applied to digital patterns, with the attacker either an
adversarial {\it source} of patterns directly
or a man-in-the-middle, intercepting a pattern on its way to the classifier.
In the former mechanism, 
there is no legitimate utility obtained from the classifier's decision.
In the latter, the decision recipient could either be the intended recipient 
or the adversary.}\label{fig:foiling}
\end{figure}

\noindent
{\bf Analysis of Test-Time Attack Mechanisms:}
Beyond the above arguments, 
we can recognize
that there are only two {\it digital} attack mechanisms on a pattern to 
be classified (see Figure \ref{fig:foiling})\footnote{Physical attacks, e.g. defacing a road sign, are not considered in this analysis.}.
In   
 one mechanism, there is
an honest generator of patterns, but the pattern is then intercepted and perturbed by an adversary 
(a ``man-in-the-middle'' attack) 
before being forwarded to the classifier.
However, there is a second mechanism, with the adversary the {\it originator} of the pattern.
Here, the adversary may have his/her own cache of labeled patterns from the domain.  
He/She selects one (even arbitrarily), perturbs it to induce a desired misclassification, and then 
sends it to the classifier\footnote{In both mechanisms, if the attacker knows the classifier's decision rule, he/she can
make a perturbation that is guaranteed to induce a change in the classifier's decision, to a target class.}.  
Only for the first mechanism, where there is a legitimate pattern, forwarded by
an honest party who is interested in the classifier's decision, is there utility in correctly classifying 
the attacked
pattern.
For the latter (adversarial source) mechanism, 
the classifier's decisions are only being made for the possible benefit
of the adversary.
Moreover, even for the man-in-the-middle mechanism, it is 
pointless to make correct decisions
if they are only going to be intercepted and modified
by the adversary on their way back to the intended decision recipient. 
(see Figure \ref{fig:foiling}).  We thus conclude that, {\it if the pattern has been attacked},
it is only meaningful to correctly classify when there is an honest generator {\it and}
when the intended recipient receives the classifier's decision.
While this may be a common scenario, an adversarial source is also a common scenario.
However, even under the man-in-the-middle scenario, for the reasons articulated previously, attack detection
in high stakes security settings is important, irrespective of making correct decisions.
If no attack is detected, one can still make a best effort to correctly
classify the pattern.
Of course, for a given application one must weigh the importance of detection
against that of (unimpeded) classification -- patterns falsely detected as attacks may be rejected (they may not be classified).  Thus, if, to achieve a desirable TPR, the FPR (equivalently, the classifier's rejection rate) is too high, attack detection may be impractical for the given domain.

\subsection{Validity of the TTE Requirement that the Attacker Knows the Ground Truth Label}

A key assumption made in all TTE attack works \cite{Szegedy_seminal},\cite{Goodfellow},\cite{Papernot},\cite{Wagner17}
is that, in addition to knowing the classifier, the attacker knows the ground-truth label of the pattern to be attacked.
This is required so that an attack perturbation can be crafted 
that pushes the pattern across the classifier's decision boundary to the decision region for another class; recall Figure \ref{fig:source-dest}.
  Without knowledge of the true label, an attacker could still perturb a pattern to ensure a {\it change} in the classifier's decision, but such change might actually alter an {\it incorrect} decision to a correct one.  Such failed attacks, which {\it correct} the classifier, could be relatively common for domains with high classifier error rates, {\it i.e.} where there is significant confusion between some classes.  Thus, in {\it all} TTE attack work it is assumed
that the true label of the pattern to be attacked is known.  Keeping this discussion in mind,
consider the man-in-the-middle attack scenario of Figure \ref{fig:foiling},
where the true label of the image to be attacked 
is actually {\it unknown} 
to the attacker!  That is, a universal assumption in TTE attacks 
is in fact not valid in an important (man-in-the-middle) scenario.

\subsection{Underestimating Susceptibility to AD in AL Attacks}

A critical review of AL research is \cite{MLSP17}. 
One concern raised is that some attacks 
are highly susceptible to an AD defense; yet this was not considered in 
performance-evaluating the attack's ``success''. 
Consider the JSMA TTE attack from \cite{Papernot}.
While the perturbed patterns in \cite{Papernot} do alter the classifier's 
decision, 
they also manifest artifacts (e.g. salt and pepper noise)
quite visible in their published figures (see e.g. 
Figure \ref{fig:JSMA-matrix}),
i.e., they are human perceptible, at least for MNIST.
While the professed requirement in \cite{Papernot} that the attack be {\it human-imperceptible} may not be so important
in practice (this is discussed further, shortly), the attack should still not
be readily {\it machine-detectable}, i.e. by an AD.
The artifacts in \cite{Papernot} are easy to
(automatically) detect in practice.
ADA  achieves an ROC AUC greater than 0.99 for this attack on MNIST digits 
and an impressive 0.97 AUC is even achieved by a simple region-counting based AD 
\cite{MLSP18-ADA}\footnote{Clean MNIST digits typically consist of a {\it single}
white region (where a region is defined as a collection of pixels that are ``connected'', with two white
pixels connected if they are in the same first-order (8-pixel) neighborhood, and with the region defined
by applying transitive closure over the whole image).
By contrast, nearly all JSMA attack images have extra
isolated white regions (associated with salt and pepper noise).  Simply using the number of white regions
in the image as a decision statistic yields 0.97 AUC on MNIST -- this strong result
for this very simple detector indicates the susceptibility of the JSMA attack to a (simple) AD
strategy.}.  Thus, at least for MNIST, JSMA attack is highly susceptible to AD.

As a second example, consider the RE attack in \cite{Reiter}.
The authors demonstrate that, with a relatively modest number of queries to the true classifier (perhaps as many as ten thousand or more,
used to create labeled training examples), one can learn
a surrogate classifier for the given domain that closely mimics the true classifier.  
Significantly, 
RE attacks may be critically needed to {\it enable} TTEs, by providing the required knowledge
of the classifier to the TTE attacker if this is unknown {\it a priori}.
One weakness of \cite{Reiter} mentioned earlier is that it does not consider large classification domains or DNNs. 
However, a much more critical weakness stems from one of its (purported) greatest
advantages -- the reverse-engineering in \cite{Reiter} does not require {\it any} real training
samples from the domain\footnote{For certain sensitive domains, or ones where obtaining real examples is expensive, the attacker may in fact have no realistic means of obtaining a significant number of real data examples from the domain.}.  In fact, in \cite{Reiter}, the attacker's queries to the classifier 
are {\it randomly} drawn, e.g. uniformly, over the given feature space.
What was not recognized in \cite{Reiter} is that
this makes the attack easily detectable -- randomly
selected query patterns
are very likely to be extreme outliers, of all the classes.  Each such query is thus
individually highly suspicious.  Multiple queries
are thus easily detected as jointly improbable under a null distribution (estimable from the training set defined over all the classes from the domain).  Even if the attacker employed {\it bots}, each
making a small number of queries, each bot's queries should also easily be detected
as anomalous. Later, we will evaluate a variant of the ADA TTE defense \cite{ICASSP19} against the more evasive RE attack proposed in \cite{Papernot3} and show that it is
quite effective (and, thus, effective at thwarting RE-enabled TTE attacks).  We further note that even 
more recent work on black box TTE attacks \cite{Madry_bandits}, which is also based on querying to estimate gradients, ignores
the possibility that queries may be detected as anomalous by an AD (a possibility verified by \cite{ICASSP19}). 

\begin{figure}[h]\centering
\includegraphics[width=0.75\columnwidth]{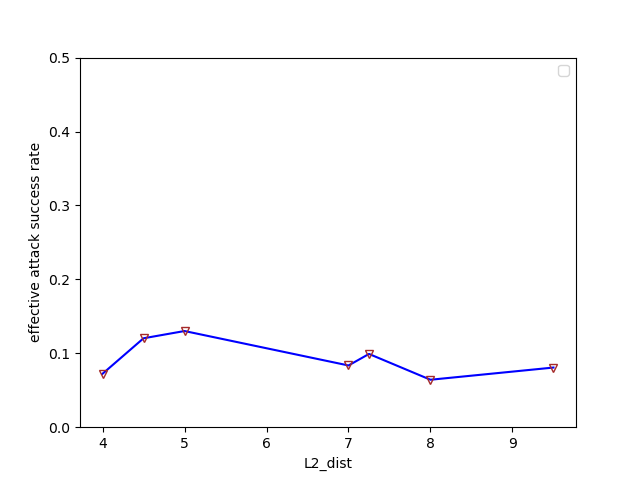}
\caption{Effective attack success rate versus attack strength for the CW attack and the ADA detector defense on the CIFAR-10 domain.}\label{fig:effective}
\vspace{-0.1in}
\end{figure}

Finally, consider evaluation of classifier performance versus attack strength in
\cite{Biggio_wild}.   
The authors give illustrative (fictitious)
curves, showing how the rate at which TTE attacks successfully cause misclassifications should increase monotonically with the strength of the 
attack (i.e., the maximum allowed size of
the perturbation).  This curve
(and \cite{Biggio_wild} in general) ignores the effect of an AD defense --
as the attack perturbation size is increased, it is indeed more likely that an attack pushes an example (say from class $c_s$) over the classifier's decision boundary (to the decision region of another class, $c_d$).
However, the attacked example is also more
easily detected (by an AD) as an outlier, with respect to (null density models for) {\it all} classes.  This 
has been experimentally
validated in \cite{MLSP18-ADA}.  
Moreover, the {\it effective} attacker success rate (the rate of crafting of samples that are both 
misclassified {\it and} not detected) may have a peak at a finite attack strength.  This is shown in Figure \ref{fig:effective} for the ADA detector and the CW attack on the CIFAR-10 domain.  Note also that the peak effective
attack success rate is about 13\%.
Thus, if an AD is in play, the attacker's ultimate success
rate does {\it not} generally strictly increase with the size of the attack.  This may also be the case for DP 
attacks, unlike the picture presented in \cite{Biggio_wild} -- as the amount of injected poisoned
training data is increased, the learned classifier's generalization accuracy should indeed degrade.  However, too much data poisoning could make the attack more easily {\it detected}.  For example, in the case of a backdoor attack, one
injects patterns mislabeled to class $c_d$ so that the classifier learns
a ``backdoor''.  
As the number of injected samples increases, there may be very distinct clusters of samples labeled to $c_d$, one corresponding to
genuine samples from $c_d$ and the other to the attack.  A clustering algorithm may
identify these two clusters as the attack strength (number of injections) is increased. 
Experiments involving
defenses against backdoor DP attacks will be given in Section V. 

\subsection{Small Perturbations for TTE Attacks: Fallacy or Requirement?}

Another fundamentally important issue is the ``proper'' attacker formulation 
and assumptions for 
TTEs.  The early approaches from \cite{Szegedy_seminal}, \cite{Goodfellow}, and \cite{Papernot}
all involve constrained optimization, with the attacker seeking to {\it minimally} perturb a test pattern
so as to induce a misclassification.  Some nuance is required here in discussing the motivation for
``small'' perturbations.  \cite{Papernot} and even more recently \cite{Goodfellow_traitor} 
specifically motivated small perturbations 
so that the attack is not {\it human-perceptible}.  However, in some application scenarios,
human monitoring of signals (images, audio) may be infrequent, 
or even non-existent.  In such cases, one might imagine that
larger perturbations, with greater success rate in causing misclassifications, could be invoked.  However,
as just noted, larger perturbations increase the {\it detectability} of the attack.  Thus, so long as an AD 
is deployed along with the classifier, ``small'' perturbations should be needed in general, to defeat machine detection (AD), irrespective of human perception.
In \cite{Biggio_wild}, on the other hand, the authors argue that it is a misconception that
``adversarial examples should be minimally perturbed''.  Interestingly, the reason given is that ``the
attacker will aim to maximize the classifier's confidence on the desired output class, rather than
only minimally perturbing the attack samples.'' Clearly, \cite{Biggio_wild}
is assuming that, {\it even if a detector is deployed}, its decisionmaking is solely based on the classifier's
``confidence'' in its decisionmaking (with attacks detected only on low confidence examples).  However, this
is a weak detector.  A well-designed AD, one which looks for {\it signatures} of an attacked
image (in \cite{MLSP18-ADA}, such signatures are anomalies, with respect to class-conditional
null densities, evaluated on the deep hidden 
layer activations of the DNN), should exhibit greater detection power as the 
attack strength is increased.  Indeed, this is what is shown in \cite{MLSP18-ADA}.  Thus,
in contrast to \cite{Biggio_wild}, we argue (and show in Figure \ref{fig:effective}) that in many application domains ``small perturbations''
are indeed a requirement, not a fallacy.

\subsection{Transferability in TTE Attacks: Targeted vs. Non-targeted cases}

\cite{Papernot3} considers the case where 
the TTE attacker does not have knowledge of the classifier but has the ability to query it.  Such querying is used to create
a labeled training set for learning a {\it surrogate} classifier.  The authors demonstrate that TTE attack patterns crafted to induce misclassification by the surrogate classifier with high success rate {\it also} induce misclassifications on the true
classifier.  Moreover, the authors demonstrate that such {\it transferability} of the TTE attack does not even require that
the true classifier and its surrogate have the same structure  -- if the true classifier is a DNN, the surrogate could be an SVM or a decision tree.  While these results are quite intriguing, it is also important
to note that the results in \cite{Papernot3} assume that the TTE attack is {\it untargeted} -- {\it i.e.}, successful transferability only requires that a misclassified TTE pattern for the surrogate is also a misclassified pattern for the true classifier.  However, for TTE attacks to be strategic and cause the most damage, the attack should be {\it targeted}, inducing assignment to a particular target class $c_t$ when the original pattern comes from class $c_s$.  In the targeted case, a TTE attack only successfully transfers
if the perturbed pattern induces the true classifier to misclassify to $c_t$.  Considering this targeted attack case, we have experimentally evaluated the RE approach in \cite{Papernot3}.  For 
CIFAR-100, using this approach to learn an 8-layer ResNet classifier that is a surrogate of the ResNet-16 true classifier resulted in a targeted transferability success rate for the CW attack of 56\%.  By contrast, the non-targeted transferability success rate was found to be over 80\%.  
This indicates at any rate that
the targeted case is more challenging, with a significantly lower success rate, than the non-targeted case.
\subsection{Black or White Box TTE Attacks for Evaluation of a Defense}
Very different viewpoints have been put forth on what is required in evaluating a defense
(even to meet the standard of publication).  Discussing TTE attacks, \cite{Wagner17} has asserted
(without elaboration) the {\it strong white box requirement} that ``one must also show that an adversary aware of the defense [white box]
can not generate attacks that evade detection.''  
There are several points to make about this.  First, it does not mention
false positives -- there is always a true detection/false detection tradeoff, with the FPR controlled by
setting the detection threshold.  Clearly, the FPR must be sufficiently low for the deployed system to be
practicable (not generating too many false alarms).  Thus, we can safely construe the requirement from
\cite{Wagner17} to be that the true detection rate should be nearly perfect, with the FPR acceptably low.
One might speciously argue in support of such a requirement by appeal to {\it worst case}
engineering analysis -- {\it i.e.}, any successful attack could have worst-case {\it consequences}, which
should be avoided at all cost.  However, it is also possible, in this highly asymmetric, fully white box 
setting, where the attacker knows everything
(the classifier {\it and} the detector, including even the detector's threshold value)
and the (proactive) defender knows nothing about
the attacker, that meeting the requirements of \cite{Wagner17} is theoretically impossible.  In fact,
\cite{Magnet} is not very sanguine about the prospects for defense against a fully white box attack. 

To put this into further perspective, consider a {\it robust classification} defense against a white box attack.
Since the attacker has full knowledge of the robust classifier, he/she may nearly always be successful in crafting a perturbed 
pattern that will induce a desired misclassification.  The only defense ``safety net'' for a robust classifier in this case is human monitoring -- if, to defeat the robust classifier, the attack perturbation is large, the attack may be human-perceptible.  However, as noted before, one cannot rely on human monitoring,
as it may be infrequent.
In this case, robust classification offers {\it no} protection against white box attacks. 

Two fundamental questions need to be asked at this point:
1) In what scenarios is a white box attacker actually realistic?  2) What should be {\it realistically expected}, performance-wise,
for a proactive (AD) defense against a white box attack?

Addressing the first question, we note that \cite{Wagner17} and other works do not justify or motivate the white box
attacker assumption.  We believe 
that, at least in security-sensitive settings where the
deployed defense will {\it not} be common knowledge, ``omnipotent'' white box knowledge of both the classifier {\it and} the AD defense would have to come from a well-placed ``insider'',
working with the attacker.  
It is {\it not} reasonable to expect an AD system to defeat an insider attack.  There are defenses specifically designed to prevent or to detect insider threats, with the former based on system 
safeguards (e.g., ``need to know'' knowledge compartmentalization) and with the latter exploiting information {\it well beyond} just the
attacked data -- {\it e.g.} {\it monitoring} suspicious behavior of users with sensitive access to the 
defense system.  Thus, we believe a fully white box attack, consistent with an insider threat, should be evaluated against a ``holistic'' defense that includes both insider threat detection/prevention and AD-based attack detection.  
Moreover, even assuming insider knowledge, true white box knowledge is in fact not in general possible, as one can always
introduce {\it randomness} into the defense that even an insider cannot predict.  For example, suppose several 
distinct TTE defenses are mounted, and for each presented test pattern the invoked defense is randomly chosen from the mounted set (e.g., using the CPU clock 
for random number generation).  In this case, the attacker will not know which defense is being used against the crafted TTE
attack pattern.  He/she can mount multiple attacked patterns, to increase the possibility of matching an attack to a defense
realization, but this will increase the attacker's work factor. 

Addressing the second question, we expect a good
defense to fundamentally alter the picture given in \cite{Biggio_wild}, where the attack success rate grows monotonically with the attack strength -- whether the attack is black box or {\it white} box with respect
to the deployed AD defense, increasing the size of a pattern perturbation increases {\it both} the probability 
that the pattern is pushed across the classifier's decision boundary and the probability that the attack is
detected.  A successful attack requires inducing a misclassification {\it and} evading detection.  Thus,
it is quite possible, with an AD defense in play, that the {\it effective} attack success rate (the joint probability of these two events) does {\it not}
grow monotonically with attack strength.  This was demonstrated in Figure \ref{fig:effective} for black box attacks (ones that know the classifier but not the detector).  In \cite{MLSP18-ADA},
this phenomenon was also demonstrated for a fully white box attack -- there was
a noticeable, single {\it peak} effective attack success rate at an intermediate attack strength.
In particular, \cite{MLSP18-ADA} devised
a white box version of the CW attack \cite{CW} on the ADA defense applied to CIFAR-10 image classification, and found, at its peak, $\sim$ 
25\% of crafted attack images induce misclassifications {\it and} evade detection. That is, 
since the induced misclassification rate is close to 100\%, roughly 75\%
of successfully {\it crafted} attacks (those causing misclassifications) are detected, at the attack strength yielding highest effective success rate.  Clearly, at the optimal attack strength, the proactive ADA defense
is not ``defeating'' the white box attack.  However, it is increasing the work factor of the attacker in producing successfully attacked images.  In fact, as reported in \cite{MLSP18-ADA}, the attacker's computational effort (to induce misclassifications) is quite high even in the absence of an AD defense -- the effort to craft CW attacks is hundreds of times heavier than that required to make ADA-based detections \cite{MLSP18-ADA}, {\it cf.} Section \ref{sec:expt}.  To defeat both the classifier and the detector, the white box attacker's computational effort is made even greater.

While \cite{Wagner17} requires strong detectability of white box attacks, other opinions on proper defense evaluation have been offered.
In \cite{Goodfellow_traitor}, the authors
quote \cite{Tygar11} which states that ``if the model gives the adversary an unrealistic degree of information...
an omnipotent adversary who knows [everything] can...design optimal attacks... it is necessary to carefully
consider what a realistic adversary can know.''

One such realistic scenario is where the attacker has full knowledge of the classifier but
no knowledge of a possible AD defense (white box w.r.t. the classifier and black box w.r.t. the detector).
This scenario is not only preferred because it is far less asymmetric ({\it i.e.}, fairer) than the
strong white box attack scenario.  It is also arguably quite realistic.
In particular, for many application settings, the best-performing {\it type} of classifier ({\it e.g.}, a DNN or a support vector machine) may in fact be common knowledge.  Moreover, even if the classifier's parameters are unknown, if the attacker has good labeled data from the domain, he/she can train a surrogate classifier with similar performance as the target classifier.  
Again, it has been shown that using surrogate classifiers
as the basis for TTE attack crafting may yield a high rate of success {\it transferability} to the target classifier \cite{Papernot3}.  Moreover, even for domains where transferability is not so high, an RE attack, involving relatively numerous queries to the target classifier, can be used to learn a good surrogate of the target classifier.  Thus, it is plausible to start with good knowledge of the classifier in play, and 
even if one does not start with such knowledge, there is a direct mechanism ({\it querying}) that allows the attacker to learn the classifier.    
However, the same {\it cannot} be said for the detector -- there may be no natural mechanism for ``querying'' the detector\footnote{If the attacker might receive the classifier's decision, one should not output a ``don't know'' decision when attacks are detected, as this would in fact allow detector querying by the attacker.}.  Moreover, with numerous potential defense approaches and no definitive standard, it seems less realistic for the attacker to possess substantial knowledge of the deployed AD defense.  Also, no one has shown there is high transferability of a surrogate {\it defense}, {\it i.e.} a {\it surrogate} AD deployed by the attacker may not closely mirror detections made by the true deployed AD.
Based on the above arguments, a reasonable framework may involve a proactive detector and an attacker that is {\it white box} with respect to the classifier and {\it black box} with respect to the detector.

\section{Attack-Defense Experiments}\label{sec:expt}
\subsection{Test-Time Evasion Experiments}
We experimented on the CIFAR-10 and CIFAR-100 \cite{cifar10} data sets.
CIFAR-10 is a ten-class data set with 60,000 color images, consisting of various animal and vehicle categories.
CIFAR-100 is a 100-class data set with 60,000 color images.
Both data sets consist of 50,000 training images and 10,000 test images, with all classes equally represented
in both the training and test sets.  For AD purposes, the data batch under consideration in
our experiments consists of the test images plus the crafted attack images.
For training DNNs, we used mini-batch gradient descent with a cross entropy loss function
and a mini-batch size of 256 samples.
For CIFAR-10, we used the 16-layer DNN architecture in \cite{He}, \cite{CW}. This trained DNN achieves an accuracy of 89.47\% on the CIFAR-10 test set.
For CIFAR-100, we used the ResNet-18 architecture, which reaches an accuracy of 81.73\% on the test set.
During the phase of crafting adversarial samples, we only perturbed test set samples that were correctly classified.
We implemented the fast gradient step method (FGSM) attack \cite{Goodfellow} and used the authors'
supplied code for the CW-L2 attack \cite{CW}.
For each test sample, from a particular class, we randomly selected one of the other classes as target
and generated an attack instance starting from the test image,
such that the classifier will assign the perturbed
image to the target class.
In this way, 
on CIFAR-10, for the FGSM attack,
we successfully crafted 9243 adversarial images (step size for gradient descent of 0.001) and
for the CW attack, we successfully crafted 9920 images (Lagrange multiplier $c$ set to 4).
Below in Figure \ref{fig:CW-array}, for visual assessment, we show successful CW attack images for the MNIST domain.  Note that while CW is generally thought of as an ``imperceptible'' attack, there
are some ghost artifacts, especially horizontally oriented, that are noticeable (with similar ghost artifacts produced by FGSM).
For CIFAR-10, though, we did indeed find CW and FGSM attacks to be
visually imperceptible.
\begin{figure}[h]\centering
\vspace{-0.15in}
\includegraphics[width=\columnwidth]{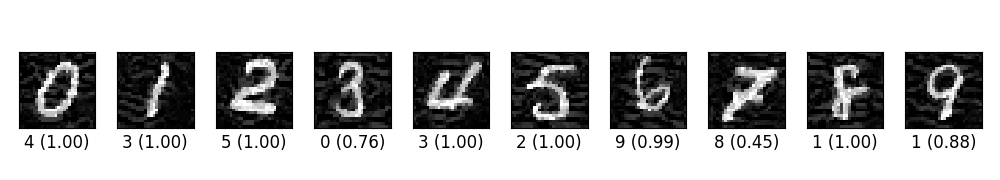}
\caption{CWL2 attack adversarial images \cite{CW} on MNIST.  Below each image the digit is the classifier's decision and the number in parentheses is the posterior probability under this winning class, indicating the decision confidence.}\label{fig:CW-array}
\vspace{-0.1in}
\end{figure}

We implemented the unsupervised AD from \cite{MLSP18-ADA} (referred to as L-AWA-ADA) and the blurring ``correction'' approach
from \cite{Li_ICCV}.  We used the code provided by the authors for the Openmax detector \cite{Openmax}.  
As mentioned earlier, Openmax requires setting two hyperparameters using labeled attack examples.  Thus, this
method is actually supervised.

\begin{figure}[h]
\begin{tabular}{c}
\includegraphics[width=0.5\columnwidth]{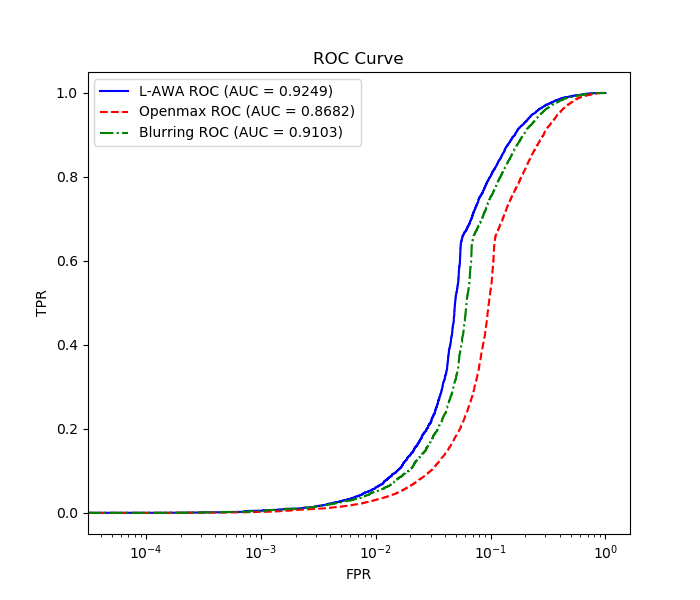}\\
\includegraphics[width=0.5\columnwidth]{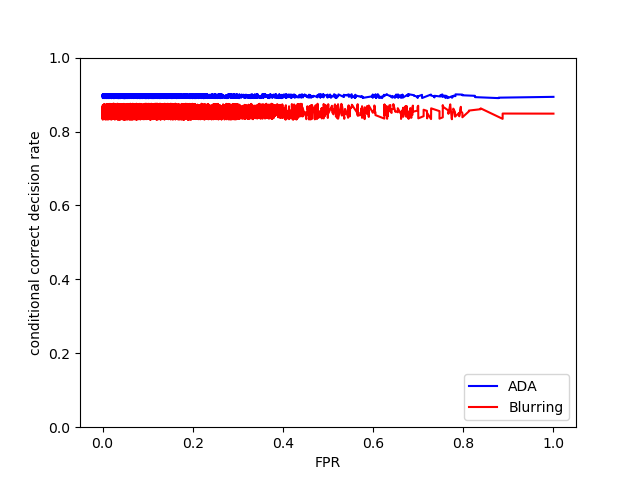}
\end{tabular}
\caption{Detector comparison for the FGSM attack on CIFAR-10.
The conditional correct decision rate (bottom) is the
correct decision rate of the classifier on the non-attacked test patterns that are not detected as attacks.
}\label{fig:compare-FGSM-CIFAR}
\end{figure}

Figure \ref{fig:compare-FGSM-CIFAR}
shows the performance of the three detectors against the FGSM attack on CIFAR-10 \cite{MLSP18-ADA}. In this case, we used 137 adversarial images to determine the hyperparameters for the Openmax method. We can see that although the blurring method's detection performance is somewhat competitive with ADA \cite{MLSP18-ADA}, this is achieved with some compromise in the (conditional) classification accuracy as the attack FPR is varied.
Openmax's detection accuracy is the worst, despite its use of (quite a few) labeled attack examples
for setting its hyperparameters.

We also assessed the execution time required by methods to make detections.
Currently we are using two NIVIDA GTX1080 GPUs, an Intel Core i7-5930K Processor and a 256GB DDR4. For this platform, detection time for a CIFAR-10 image is 0.00114s using ADA and 0.00063s for the blurring method, {\it i.e.} it is both quite modest and
comparable for the two methods.  For comparison's sake, the average time required by the attacker to craft a successful CW attack example
on CIFAR-10 (averaged over the whole test set) is 0.72s -- about 700 times that required for ADA detection.
Thus, detector computational complexity is much lower than that of the attacker.

\begin{figure}[h]\centering
\vspace{-0.15in}
\centering\includegraphics[width=0.75\columnwidth]{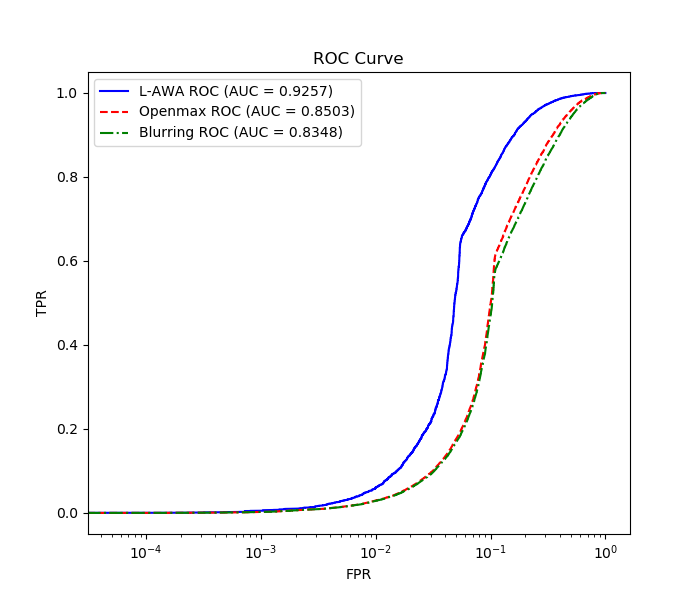}
\caption{Detector comparison for the CW L2 attack on CIFAR-10.  
}\label{fig:compare-CWL2-CIFAR}
\vspace{-0.1in}
\end{figure}

For the CW L2 attack
(Figure \ref{fig:compare-CWL2-CIFAR}),
we used 152 adversarial samples to determine the  hyper-parameters for Openmax.
ADA greatly outperforms the other detectors in this experiment.
Again, although not shown, the blurring method compromises some classification accuracy
as the FPR is varied.

\begin{figure}[!htb]
\includegraphics[width=0.75\columnwidth]{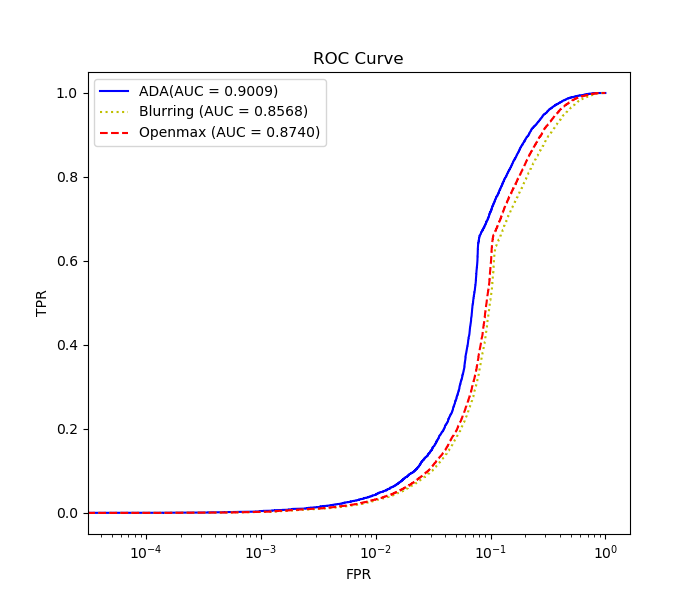}
\caption{CW attack detection results on CIFAR-100.}
\label{cifar100}
\end{figure}

We also evaluated the methods on CIFAR-100 \cite{MLSP18-ADA}.
The only modification of the ADA method for this data set was not to perform inverse
weighting by the confusion matrix, as the confusion matrix is not reliably estimated
in the 100-class (limited training set) case.  Again, OpenMax used 150 labeled attack examples to choose its hyperparameters.  Figure \ref{cifar100} 
shows that ADA significantly outperforms the other detectors.
Moreover, while ADA's ROC AUC is lower than for CIFAR-10 (as might be expected), it is
still above 0.9.

\begin{figure}[!htb]
	\includegraphics[width=0.75\columnwidth]{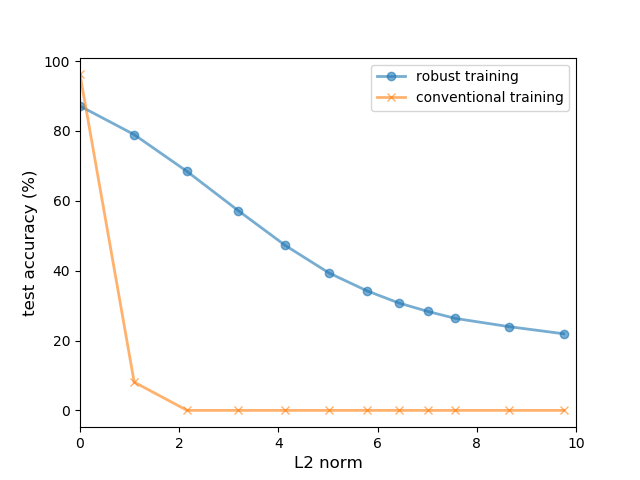}
	\caption{Test accuracy (\%) of PGD adversarial samples for an adversarially trained DNN and a conventionally trained DNN, over a range of attack strengths.}
	\label{fig:robust_training}
\end{figure}

In \cite{MLSP18-ADA} ADA was also demonstrated to substantially outperform the unsupervised detection method from \cite{AD};  moreover, ADA's results on the CW attack on CIFAR-10 were compared with the results
for a variety of detection methods reported in \cite{Wagner17} and found to be better than the 
best results reported in \cite{Wagner17}.  While the ADA approach does give relatively promising results in detection of TTE attacks, a ROC AUC of 0.9 is far from perfect -- there is potentially room for improvement from a more powerful detector, as well as the need to theoretically characterize what is attainable.  In particular, while it is known that imperfect classification accuracy impacts detection accuracy \cite{MLSP18-ADA} (misclassified patterns are more difficult to distinguish from attacked patterns than correctly classified patterns), there is no theoretical characterization of how the classifier's error rate bounds the achievable accuracy of a detector.

In addition, we evaluated one of the state-of-the-art robust classification approaches, based on adversarial training, proposed in \cite{Madry17}. We compared its robustness against a recent powerful TTE attack, projected gradient descent (PGD) \cite{KGB16}. A conventional DNN and the robust DNN were trained on CIFAR-10 by the authors of \cite{Madry17}. The details about the training configurations and attack devising are provided in \cite{Madry17}. We launched PGD attacks (assuming the DNN parameters are known to the attacker) for a range of attack strengths against both DNNs. The attack strength of PGD attacks is controlled by the $L_{\infty}$ norm of the adversarial perturbation. For easier comparison with Figure \ref{fig:effective} which shows the {\it effective} attack success rate of CW over a range of attack strengths when the ADA detector is in play, we used the L2 norm (of the perturbation) to measure the attack strength of PGD. In Figure \ref{fig:robust_training}, we plot the test accuracy, i.e. the percentage of adversarially perturbed patterns being correctly classified, for both DNNs, for the range of attack strengths, and observe a clear gain from using the adversarially trained robust DNN over using the conventionally trained DNN. However, when there is no attack (i.e. the L2 norm equals 0), there is a significant ($\sim$10\%) degradation in accuracy for the adversarially trained DNN. Also, when the attack strength grows a little, even this state-of-the-art robust classifier suffers from great loss. For example, when the L2 norm of the perturbation is larger than 4 (or the corresponding $L_{\infty}$ norm is larger than 8), the test accuracy for the robust classifier drops below 50\% -- a detector is still in great need (and, at least for CW, ADA detects most attacks at an L2 norm of 4), as suggested by the results in Figure \ref{fig:effective}.

\subsection{Reverse Engineering Experiments}
We investigated detection of the RE attack from 
\cite{Papernot3}, which involves stage-wise querying and surrogate classifier re-training, over a 
sequence of stages, as discussed in Section II.C.  The premise of the RE detection method in \cite{ICASSP19}
is that, in order to learn about the classifier, the attacker must generate ``exploratory'' queries that may be atypical of legitimate patterns from the domain, just as are TTE attack patterns. 
Thus, \cite{ICASSP19} leverages the ADA detector (designed to detect TTE attacks) to {\it jointly} exploit batches of images from a common querying stage, seeking to detect RE attacks.  
To
aggregate
ADA decision statistics over a batch of (query) images from a given querying stage, \cite{ICASSP19} performs the following: 
i) Divide the batch into mini-batches, for example a batch of 50 images could be divided into mini-batches of size 5; ii) For each mini-batch, 
maximize the ADA decision statistic (produced for each image) over all images in the mini-batch; iii) 
Make a detection if {\it any} of the mini-batches yields a detection statistic greater than the threshold.

\begin{figure}[h]\centering
\centering\includegraphics[width=0.75\columnwidth]{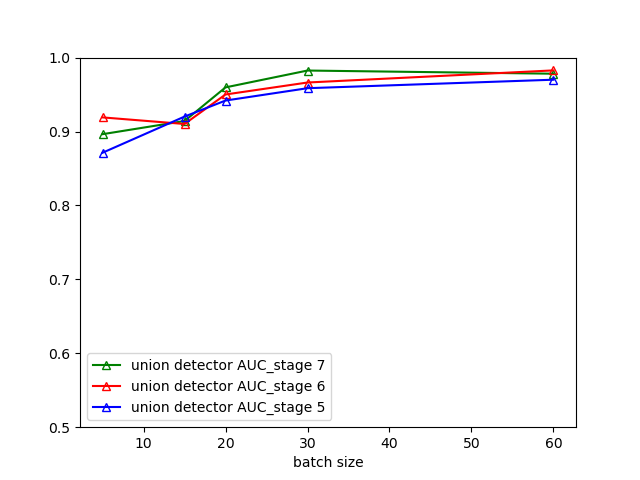}
\caption{RE detection ROC AUC on CIFAR-10 at different query stages versus batch size.}\label{fig:1}
\end{figure}

\begin{figure}[h]\centering
\centering\includegraphics[width=0.75\columnwidth]{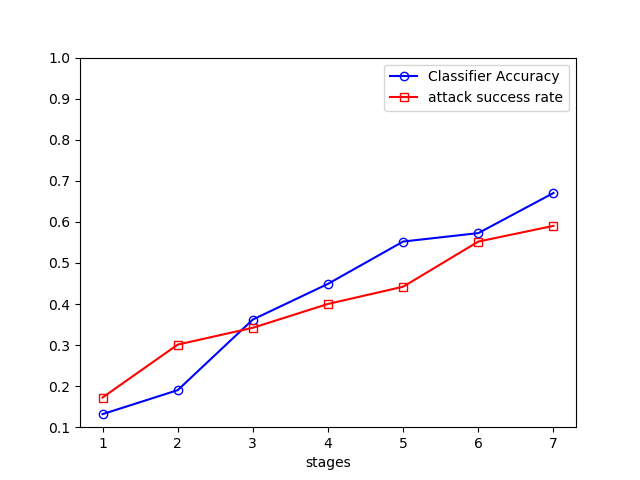}
\caption{TTE attack success rate and surrogate classifier accuracy on CIFAR-10 versus RE query stage.}\label{fig:2}
\end{figure}

For CIFAR-10, as a DNN classifier, we used ResNet-18. We also used the same structure for training the RE attacker's surrogate network.
For stage 0, we used 280 CIFAR-10 samples (28 from each class) to train the initial surrogate classifier. We applied 6 stages of retraining (7 training stages) of the surrogate DNN, using gradient descent with $\lambda=0.37$.  The number of queries generated by the 6 stages were: 280, 560, 1120, 2240, and 4480. We used mini-batches of size 5 in experiments. Two maxpooling layers and the penultimate layer were used in generating the ADA detection statistics. 
Figures \ref{fig:1} and \ref{fig:2} show the detector's and attacker's performances for different
query stages.  For CIFAR-10, the surrogate classifier's accuracy and the TTE attack
success rate are not very high, even after stage 7.  On the other hand, the ADA based
RE detector achieves high ROC AUC, above 0.97, as the batch size is increased, for 
query stages 6 and 7.
Thus, the approach in \cite{ICASSP19} is highly successful in detecting RE attacks, and at a stage before the attacker has learned a surrogate classifier sufficiently accurate to launch a strong TTE attack.
ADA thus provides two ways to defeat a TTE attack: i) at the precursor RE stage; ii) during an actual TTE attack.

\subsection{Embedded Backdoor Data Poisoning Experiments}
\begin{figure}[h]
\centering
\includegraphics[width=0.75\columnwidth]{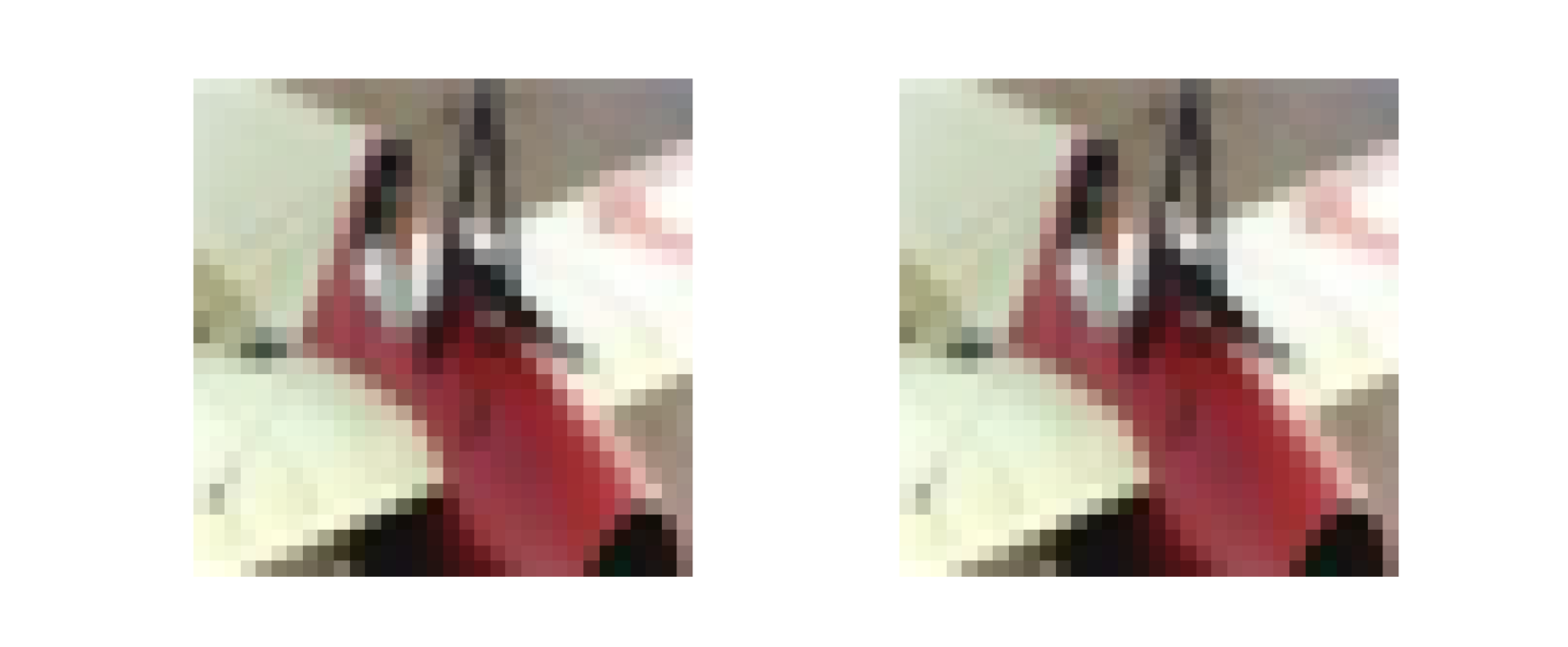}
\caption{Backdoor-attacked low-resolution image of a plane (left) with a single pixel changed from the clean image (right).}\label{fig:backdoor}
\end{figure}

We experimented on the CIFAR-10 data set with the same train-test split as in the TTE experiments. The training set is poisoned by inserting 1000 backdoor images crafted starting from clean images from the `airplane' class and then labeled as `bird'. This (source, target) class pair is one of the eight pairs used in \cite{Madry-NIPS18}. The backdoor signature is a positive perturbation (whose size is pre-selected by the attacker) on a randomly chosen pixel that is fixed through all the experiments. An example of a backdoor image (on the left) and its original clean version (on the right) are shown in Figure \ref{fig:backdoor}. It is not visually perceptible that a perturbation of size\footnote{The pixel values are normalized to [0, 1] from [0, 255] for each of the R, G, B colors.} 0.25 was added to each of the R, G, B values of the pixel at position (9, 18) (the image size is 32$\times$32). Scenarios where adversarial images are crafted using other methods, including fixing values of some pixels to form a backdoor shape (pixel, `X', or `L' shapes) with {\it e.g.} (R, G, B) = (0.1, 0.1, 0.9)), are tested in \cite{Madry-NIPS18}. However, compared with our
pixel perturbation backdoor, the backdoor signatures in \cite{Madry-NIPS18} are visually more perceptible and easier to detect in experiments. Thus, in this sub-section, we focus on perturbation attacks on a single pixel, varying the perturbation size, and compare the performance of three defenses. Note that we control the attack strength by varying the perturbation size while fixing the number of backdoor training patterns (at 1000).

\begin{figure}[h]
\centering
\includegraphics[width=0.75\columnwidth]{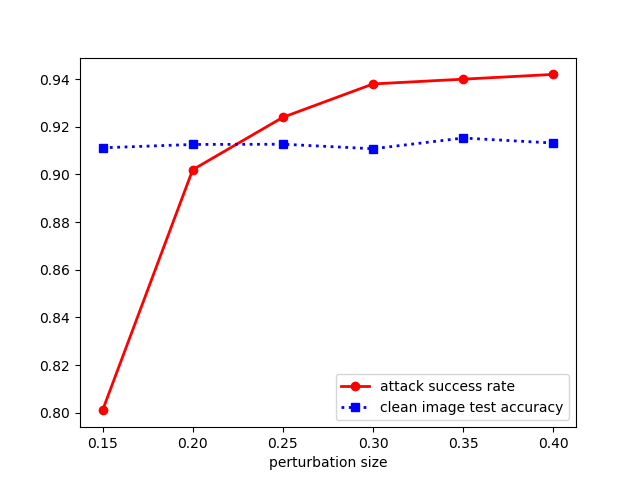}
\caption{Attack success rate and accuracy on the clean test set for a range of perturbation sizes.}\label{fig:DP_before}
\label{before}
\end{figure}

For training DNNs, we used ResNet-20 \cite{He} and trained for 200 epochs with mini-batch size of 32, which achieved an accuracy of 91.18\% on the clean test set. Under attacks with different perturbation sizes, the accuracies on the clean test patterns and the attack success rates are shown in Figure \ref{fig:DP_before}. Here, the attack success rate is the fraction of backdoor test patterns (those not used during training) that are classified to the 
target class. In our experiments, we created the backdoor test patterns by adding the same perturbation to the 1000 clean test patterns from a given class.  Again, the attacker's goal is for the classifier to assign
these patterns, originating {\it e.g.} from the `airplane' category, to the target category, {\it e.g.} `bird'. For all the perturbation sizes tested, the classification accuracies on the clean test patterns are not appreciably degraded, and the attack success rates are all quite high.

For defenses against this attack, we assess the approach based on a spectral signature (SS) proposed in \cite{Madry-NIPS18}, the activation clustering (AC) approach proposed in \cite{BChen18}, and a cluster impurity (CI) defense \cite{MLSP19-backdoor} which takes advantage of several existing methods in the broader AL defense literature. All three methods extract, when the training patterns are applied as input, the penultimate layer features from the trained DNN. SS projects a feature vector onto the principal eigenvector of the feature vector's covariance matrix. AC projects the feature vector onto 10 independent components. CI operates on the whole feature vector, without dimensionality reduction. 

SS, as pointed out in Section III.D, relies on knowledge of the number of backdoor training patterns, unlikely to be known in practice. Moreover, \cite{Madry-NIPS18} did not provide a practical way to explicitly infer whether a class is backdoor-poisoned or not. For convenience of evaluation, we fixed the FPR of SS to 0.5 by applying the same detection threshold to all classes and then evaluated the TPR.  Note that FPR=0.5 means half of the unpoisoned training patterns
will be falsely detected and removed.  This very high FPR was necessary in order for SS to achieve significant
TPR (detecting the backdoor). 

Regarding AC, even though 
this method is successful in backdoor detection and mitigation on simpler data sets
(MNIST and LISA), it does not perform very effectively for more difficult data sets ({\it e.g.} CIFAR-10) with evasive adversarial patterns (e.g. perturbation of a single pixel). The `retraining' method (one of the three methods proposed in \cite{BChen18}) for deciding which of the two clusters obtained by K-means should be discarded is an effective way to look for the poisoned classes and clusters. 
However, this method is impractical when the number of classes is large.
Without explicitly implementing the retraining method, we present essentially ``best possible'' results for AC
by only applying it to the class that has actually been attacked {\it and} by choosing for removal the cluster from this class that actually possesses the most poisoned patterns.  That is, we assume there are {\it zero} false positive
detections associated with unattacked classes (even though this is highly optimistic) and that AC unfailingly picks the correct cluster to detect and remove.

\begin{figure}[h]
\centering
\includegraphics[width=0.75\columnwidth]{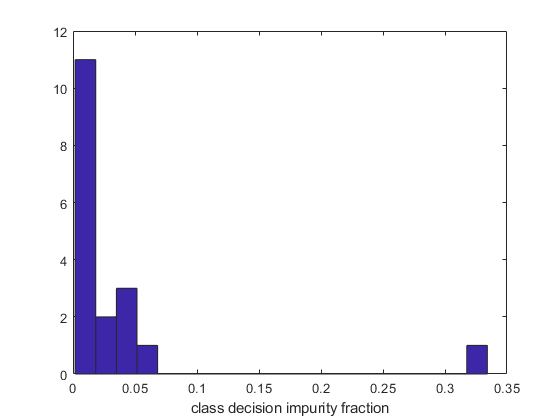}
\caption{Histogram of the class decision impurity fraction of the 18 clusters estimated using BIC for perturbation size 0.25.}\label{fig:CDIF}
\end{figure}

CI first models the feature vectors extracted from the penultimate layer for each class using Gaussian mixture models with full covariance matrices. The number of clusters for each class is selected using BIC model order selection \cite{Schwarz}. For the relatively simple MNIST data set, BIC selects one cluster if the class is not poisoned and two clusters otherwise, which agrees with the observations reported in \cite{BChen18}. However, for the more difficult CIFAR-10 data set, the BIC-optimal number of clusters is not necessarily one for the classes that are not poisoned. For example, with perturbation size 0.25, BIC selects 18 clusters in aggregate for the 10 classes, even though only one class is poisoned. Thus, we use a measured `class decision impurity fraction' for each cluster to decide whether it should be removed or not. We first apply to all training patterns the 2$\times$2 averaging filter proposed in \cite{Li_ICCV} for TTE defense.  Then each training pattern and its `blurred' version are classified by the trained DNN. The class decision impurity fraction of each cluster is then evaluated as the fraction of patterns in the cluster whose predicted labels disagree for the original and blurred versions. The backdoor signatures in poisoned patterns, once blurred, are likely to become less `recognizable' by the trained neural network, {\it i.e.} those blurred patterns are likely to be classified by the DNN to each pattern's originating class, rather than to the attacker's target class. If a cluster contains many backdoor-poisoned patterns, the class decision impurity fraction will be abnormally high.  As an example, Figure \ref{fig:CDIF} gives the histogram for the class decision impurity fraction for the 18 clusters (across all originating classes) selected using BIC, with perturbation size 0.25 (similar histograms are obtained for other perturbation sizes). 
Note the clear, large gap between low and high impurity clusters (with the single, high impurity cluster from the attacked class).
Thus, we set the impurity threshold at 0.1 for the CI approach in all our experiments.

\begin{table}
	\begin{center}
	\caption{(TPR, FPR) tradeoffs for the three detection approaches over a range of perturbation sizes.}\label{tab:DPDetection}
	\begin{tabular}{ |c|c|c|c| } 
		\hline
		Pert Size & SS & AC & CI \\ 
		\hline
		0.15: & (0.443, 0.5) & (0.622, 0.057) & (0.962, 0.002) \\
		\hline 
		0.20: & (0.483, 0.5) & (0.691, 0.040) & (0.980, 0.003) \\
		\hline 
		0.25: & (0.474, 0.5) & (0.763, 0.040) & (0.979, 0.003) \\
		\hline 
		0.30: & (0.558, 0.5) & (0.741, 0.037) & (0.976, 0.003) \\
		\hline 
		0.35: & (0.666, 0.5) & (0.851, 0.034) & (0.989, 0.002) \\
		\hline 
		0.40: & (0.616, 0.5) & (0.847, 0.036) & (0.985, 0.004) \\
		\hline
	\end{tabular}
	\end{center}
\end{table}

Table \ref{tab:DPDetection} shows (TPR, FPR) pairs of the three detection approaches for each perturbation size. In general, CI achieves much higher TPR (greater than 0.96) than the other two approaches and at the same time very low FPR (lower than 0.005).  Implemented in the most preferential way, as discussed previously, AC achieves relatively good TPR for large perturbation size. However, TPR clearly diminishes with the perturbation size. The TPR of the SS approach shares the same trend, but it is apparently not comparable to the other two approaches. Regarding FPR, we only evaluated for the AC and CI approaches since it is fixed for the SS approach. Although the FPRs of the AC approach are not very high, 
note that these numbers are quite optimistic since we assumed AC has zero false positives from non-attacked classes.
Another way of saying this is that, in fact, AC's false positive rate for (conditioned on) the attacked class is in fact quite high --
for AC, 0.036 FPR means that 1800 out of the 5000 clean patterns from the target class are falsely detected as backdoor
patterns and discarded from the training set (since zero counts are assumed for all non-attacked classes, all the false positives come from the target class) -- the FPR rate for the target class is actually 0.36.
By contrast, the FPR of the CI approach is truly low, with no more than 5\% of clean patterns falsely detected 
for {\it any} of the classes, including the attacked class.

\begin{figure}[h]
\centering
\includegraphics[width=0.75\columnwidth]{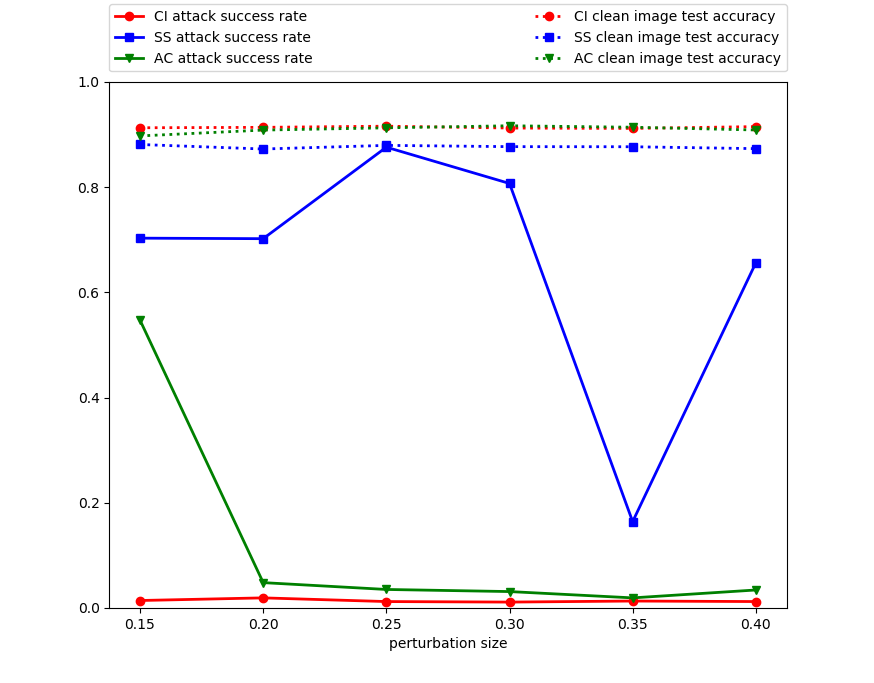}
\caption{Attack success rate and accuracy on clean test set of the retrained neural networks for the three defenses.}\label{fig:DP_after}
\end{figure}

As discussed earlier, the ultimate performance assessment for a detector-based defense (where training patterns are removed and then the classifier is retrained) involves the retrained classifier's clean test set accuracy and the 
backdoor attack success rate on the retrained classifier.
In Figure \ref{fig:DP_after} we see that the CI approach reduces the attack success rate to a negligible level, and the clean test set accuracy is not degraded at all. The AC approach is relatively successful, but its performance at low perturbation size is poor (with the attacker's success rate in the absence of defense already at greater than
80\% at this attack size, as seen from Figure \ref{before}) -- an attack success rate above 0.5 should be quite harmful. The SS approach clearly fails to defeat the attack. Moreover, there is some degradation in clean test set accuracy even though training patterns are abundant for CIFAR-10 (5000 training patterns per class). If we allow an even higher FPR to achieve a better TPR for SS, the retrained neural network will suffer even more test set accuracy degradation.

Anecdotally, different from the attacks on the simpler MNIST data set in \cite{BChen18} and those crafted using fixed abnormal pixel values in \cite{Madry-NIPS18}, the attacks used in our experiments are clearly less detectable. For the AC approach, the feature vectors corresponding to the adversarial patterns cluster together but are surrounded by clean patterns, when projected onto (a visualizable) lower-dimensional space. For the SS approach, the non-separability between adversarial and clean patterns 
is quite visible after principal component projection.
{\it One} reason for CI's advantage may be its use of all the penultimate layer features -- some ``minor component''
features may contain important information for discriminating clean from poisoned patterns.  Another likely reason is 
use of the blurring heuristic in conjunction with clustering.

Apart from the experiments above, we also applied the three defenses to the more detectable attacks in \cite{Madry-NIPS18}. 
In these experimental results  (not shown), all three methods performed quite well, obtaining good (TPR, FPR) tradeoffs.

\subsection{Backdoor Detection Without the Training Set}

\begin{figure}[htb]
	\vspace{-0.15in}
	\begin{equation*}
	\hspace{-0.05in}
	\begin{minipage}[h]{0.3\linewidth}
	\vspace{0pt}
	\includegraphics[width=\linewidth]{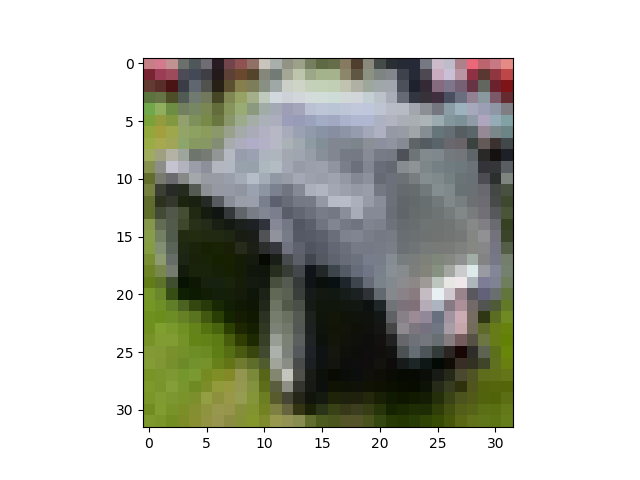}
	\end{minipage}
	\hspace{-0.05in}
	=[
	\begin{minipage}[h]{0.3\linewidth}
	\vspace{0pt}
	\includegraphics[width=\linewidth]{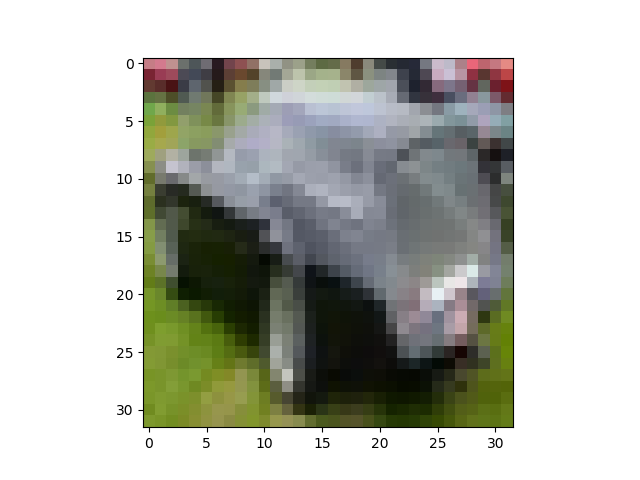}
	\end{minipage}
	\hspace{-0.05in}
	+\,\epsilon\,\times
	\begin{minipage}[h]{0.3\linewidth}
	\vspace{0pt}
	\includegraphics[width=\linewidth]{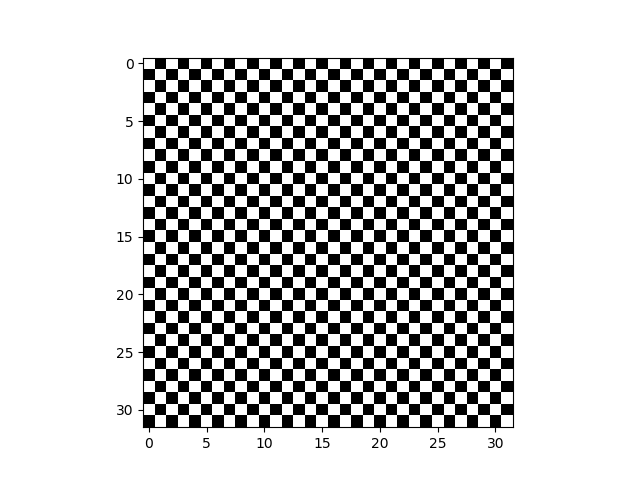}
	\end{minipage}
	\hspace{-0.05in}
	]_c
	\end{equation*}
	\vspace{-0.15in}
	\caption{\label{fig:bd_image_perceptible}
		An example backdoor image with imperceptible backdoor pattern, a ``chess board'' pattern.}
	\vspace{-0.05in}
\end{figure}

We demonstrate post-training backdoor detection of both imperceptible and innocuous, perceptible backdoor patterns. Imperceptible backdoor patterns are usually sparse pixel-wise (e.g. the one in Figure \ref{fig:backdoor}) or image-wide additive perturbations with invisibly small perturbation size/norm. Here we consider the same image-wide ``chess board'' backdoor pattern used by \cite{TrojAI} as shown in Figure \ref{fig:bd_image_perceptible}, where $\epsilon$ is the scaling factor which makes the perturbation imperceptible (e.g. $\epsilon=1/255$ was used in our experiment), and $[\cdot]_c$ is the clipping operator that guarantees each pixel value of the perturbed image is in the valid range $[0, 1]$.

\begin{figure}[!htb]
	\includegraphics[width=1\columnwidth]{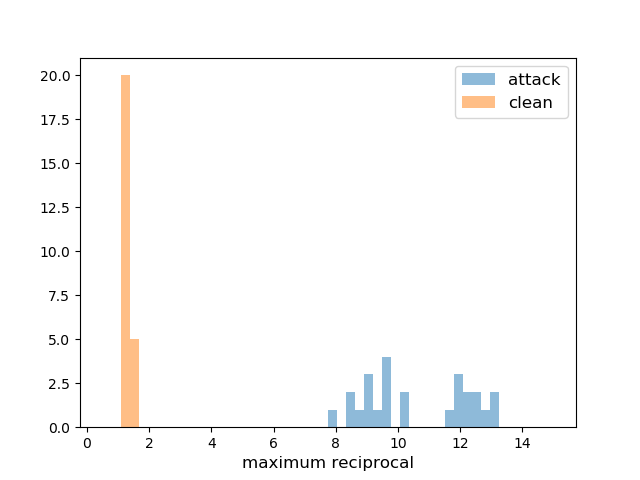}
	\caption{Histogram of the maximum reciprocal over all 90 class pairs for each of the 25 DNN realizations in the attack group and the clean group.}
	\label{fig:imperceptable_hist}
\end{figure}

\begin{figure}[t]
	\centering
	\begin{minipage}[b]{0.45\linewidth}
		\centering
		\centerline{\includegraphics[width=\linewidth]{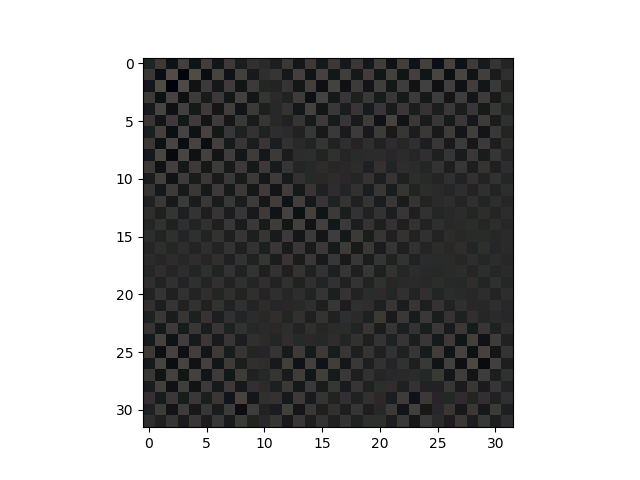}}
	\end{minipage}
	\caption{An example of the estimated perturbation (with the image scaled by a constant factor for visualization) for the true backdoor pair, which is close to the true backdoor pattern used by the attacker.}
	\label{fig:chess_board_est}
\end{figure}

We tested the state-of-the-art post-training detector for imperceptible backdoor patterns proposed in \cite{TrojAI}. This detector first estimates, using a clean, labeled data set, the minimum size perturbation required to induce a high fraction of group misclassification for all (source, target) class pairs. Then the {\it reciprocals} of the perturbation norm for all class pairs are used for detection inference. If for any class pair, the reciprocal is abnormally large, a backdoor attack is detected, with this class pair inferred to be used in devising the attack. Here we verify the premise behind this detector and show its performance using 25 realizations of DNNs being attacked and 25 realizations of clean DNNs not attacked. These DNNs are trained on CIFAR-10 with the training (and attack) configurations specified in \cite{TrojAI}. We first show in Figure \ref{fig:imperceptable_hist} the histogram of the maximum reciprocal over all 90 class pairs for each of the 25 DNNs for both the attack and the clean group. For all 25 DNNs being attacked, the maximum reciprocals are much larger than for the 25 clean DNNs. This detector achieves outstanding detection performance -- much better than an earlier detector \cite{NC}. All 25 attacks are successfully detected, among which, both source and target classes used for devising the attack are correctly inferred for 23 out of 25 attack instances; for the other two attack instances, only the target class is correctly inferred. Moreover, all 25 clean DNNs are inferred as ``not attacked''. Finally, in Figure \ref{fig:chess_board_est}, we show an example of the estimated perturbation (with the image scaled by a constant factor for visualization) for the true backdoor pair, which is very close to the true chess board pattern used for devising the attack.

\begin{figure}[htb]
	\vspace{-0.15in}
	\begin{equation*}
	\hspace{-0.15in}
	\begin{minipage}[h]{0.33\linewidth}
	\vspace{0pt}
	\includegraphics[width=\linewidth]{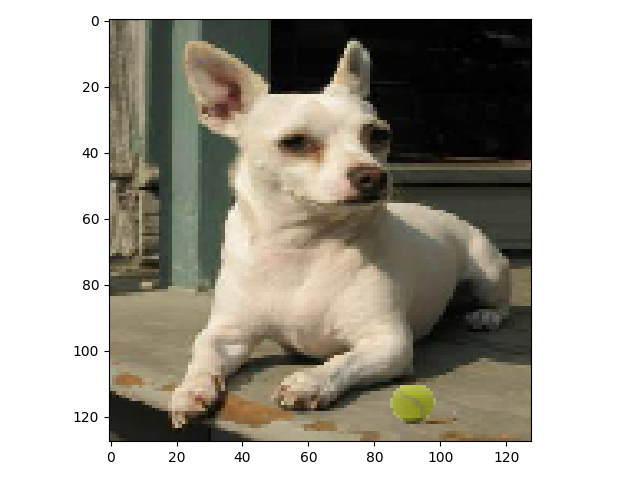}
	\end{minipage}
	\hspace{-0.15in}
	\begin{cases}
	\begin{minipage}[h]{0.33\linewidth}
	\vspace{0pt}
	\includegraphics[width=\linewidth]{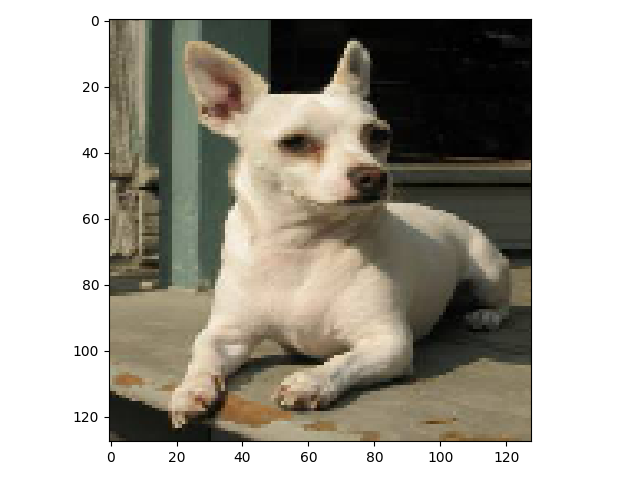}
	\end{minipage}
	\hspace{-0.1in}&\odot\hspace{+0.1in}
	\begin{minipage}[h]{0.33\linewidth}
	\vspace{0pt}
	\includegraphics[width=\linewidth]{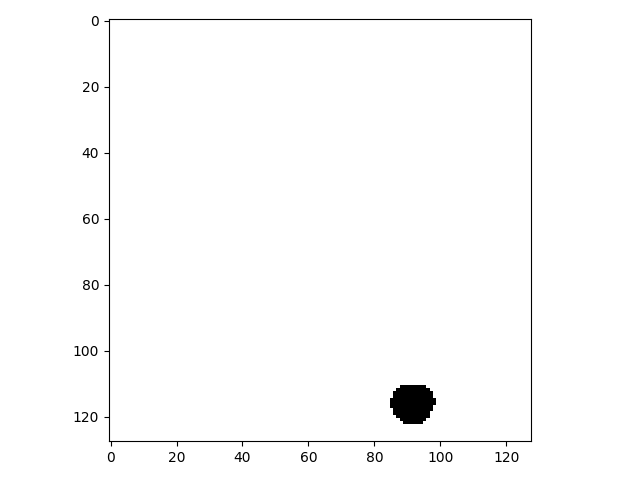}
	\end{minipage}\\
	\hspace{-0.1in}&+\\
	\begin{minipage}[h]{0.33\linewidth}
	\vspace{0pt}
	\includegraphics[width=\linewidth]{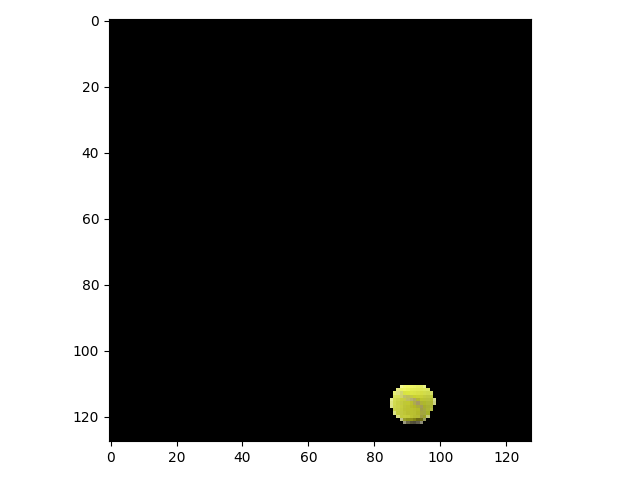}
	\end{minipage}
	\hspace{-0.1in}&\odot\hspace{+0.1in}
	\begin{minipage}[h]{0.33\linewidth}
	\vspace{0pt}
	\includegraphics[width=\linewidth]{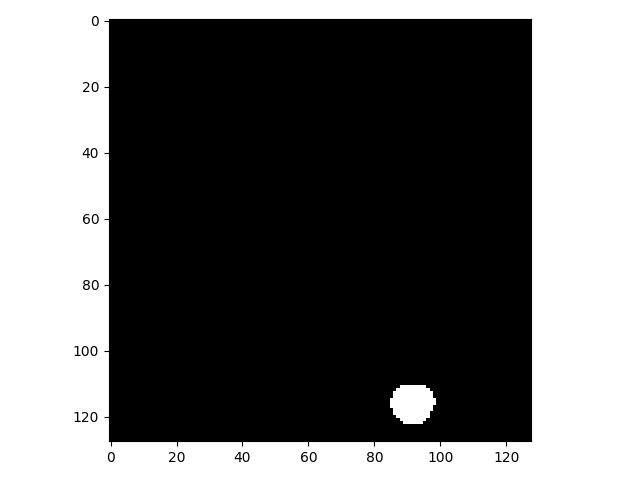}
	\end{minipage}
	\end{cases}
	\end{equation*}
	\vspace{-0.15in}
	\caption{\label{fig:bd_image_exp}
		An example backdoor image embedded with a perceptible backdoor pattern, a ``tennis ball'', created using a clean image from class ``chihuahua''.}
	\vspace{-0.05in}
\end{figure}

Perceptible backdoor patterns are usually scene-plausible objects and are embedded into clean images in a different way than imperceptible backdoor patterns. As shown in Figure \ref{fig:bd_image_exp}, a ``tennis ball'' is embedded into a clean image from class ``chihuahua''. Instead of perturbing pixels, a small region of the image is entirely wiped out by a mask, and replaced with the tennis ball.

Here we demonstrate a post-training detector designed especially for perceptible backdoors, proposed in \cite{backdoor-perceptible}. This detector is based on the fact that estimating a pattern on a spatial support with restricted size to induce a high group misclassification is usually much more achievable for backdoor class pairs than for non-backdoor class pairs. For each class pair, pattern estimation (aiming to maximize the misclassification fraction) is performed on each of a prescribed set of spatial supports with increasing size. For example, the set of spatial supports could be squares of width 5, 6, 7. For each class pair, the maximum achievable misclassification fraction (MAMF) is obtained for each of the spatial supports and averaged over all the spatial supports. If the DNN is attacked, there will be a class pair (which is likely the true backdoor pair) with abnormally large average MAMF.

\begin{figure}[!htb]
	\includegraphics[width=1\columnwidth]{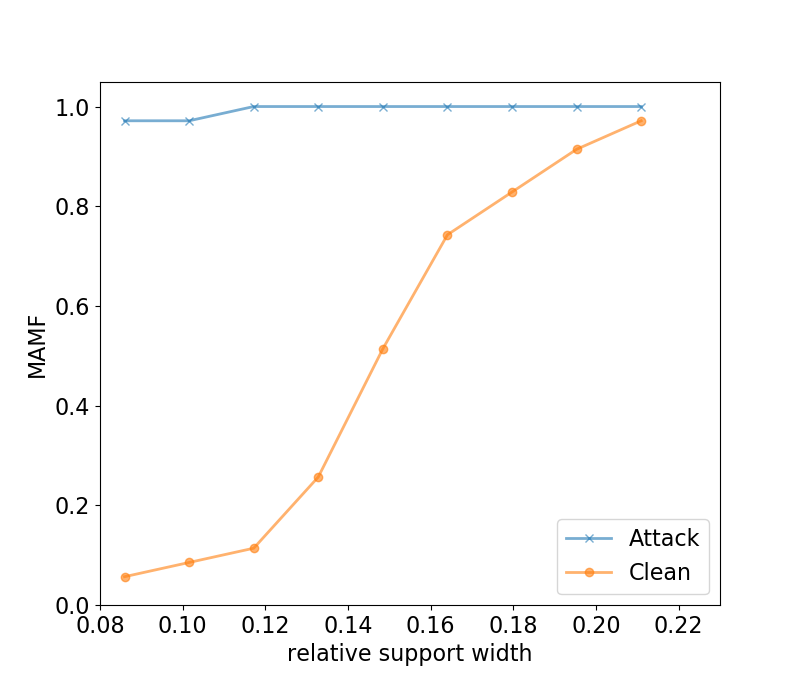}
	\caption{Maximum achievable misclassification fraction (MAMF) for the class pair having the largest average MAMF, for both the DNN being attacked and the clean benchmark DNN, considering an attack instance on Oxford-IIIT.}
	\label{fig:perceptible_curves}
\end{figure}

We consider the Oxford-IIIT data set \cite{Pets-Data} and use the same training configurations and attack settings as in \cite{backdoor-perceptible} to train a clean benchmark DNN and an attacked DNN. We also use the same detector configuration as in \cite{backdoor-perceptible}. In Figure \ref{fig:perceptible_curves}, for both DNNs, we show the MAMF over a range of relative support widths (i.e. the ratio of the actual width of the square support used for pattern estimation over the image width) for the class pair having the largest average MAMF. The clear, huge gap between the two curves shows the significant difference between the maximum average MAMF (over all class pairs) for the two DNNs. With an easily chosen threshold, the attack can be detected with high confidence.

\section{Some Continuing Directions and Challenges}\label{sec:fw}

\subsection{Combining Robust Classification with Detection}
In this paper we have emphasized that TTE attack detection is an important inference objective in its own right and one that is ignored in robust classification approaches, which seek to correctly classify despite the attack.  However, one promising possibility is to combine these two approaches, with the detector built off of the robust classifier: robust classification may force the TTE attacker to make perturbations larger to ensure misclassification success, but this may also make the attack more easily detected.  This may enable approaches like ADA to bridge their current performance gap and achieve more robust and secure classifiers.  This possibility will be investigated in future.

\subsection{Defenses with a human-in-the-loop}
Even if AD-based defenses achieved nearly perfect accuracy in detecting TTE and DP attacks (true positive rate of one,
false positive rate of zero), having a human analyst/expert in the loop would be extremely useful for characterizing
the nature of a detected attack and for determining suitable actions in response ({\it e.g.,} measures to block an attacker's
future access, potential retaliation, termination of actions in progress that may have been based on TTE attack or
a DP-compromised classifier's decisions).  Human {\it categorization} of detected attacks would also enable {\it active learning} of an automated classifier designed to mimic the human analyst, {\it i.e.}, to categorize anomalies, beyond merely detecting them.
Such an automated classifier could be very useful, considering that the volume of detected anomalies may be quite large,
whereas an analyst may only be sparingly available to inspect them, given the required effort and cost/expertise.

Beyond these benefits, consider that the results here show that {\it good}, but far from perfect, TTE attack detection is
currently achieved. Sparing human-in-the-loop labeling of detections as ``attack'' or ``no attack'' would
enable {\it partially} supervised 
(semisupervised \cite{MILLER,Qiu16}) learning of the 
anomaly detector.  Such learning could bridge the existing performance gap in detecting the most challenging
attacks, such as CW \cite{Wagner17,CW}.  For example, the ADA approach \cite{MLSP18-ADA} could be modified so as to learn a supervised
classifier that takes as derived feature {\it input} ADA's rich set of KL decision statistics -- 
the classifier could be a logistic regression model, a support vector machine, or even (ironically) a deep neural network\footnote{In the last case, a DNN is being trained to decide when a different DNN is being ``fooled''.}.
In this case, the classifier
would be learning how to weight individual feature anomalies and which patterns of anomalies are most discriminating between
TTE attack and non-attack instances. 
\subsection{The Challenge of Fake News and More General Injected Content Attacks}
\subsubsection{Sophisticated data poisoning attacks on classifiers}
We have demonstrated some potential of clustering and AD to detect DP attacks.  
However,
consider recent advances in creating realistic simulated data, such as generative adversarial networks (GANs) \cite{gans}.  As an example, suppose 
one hand-crafts a small set of 
realistic-looking images of fighter aircraft used by country A, {\it albeit}
with key but subtle characteristics altered so as to introduce ambiguity with fighters used by another country, B.
GANs \cite{gans} could be subsequently used to mass-produce many such instances, starting from the small hand-crafted set.
If these synthetic examples (labeled by A) are used in a DP attack, a learned classifier may mistake aircraft from
B with those of A, with potentially dire consequences.  These sophisticated DP attacks may be difficult to detect using clustering or outlier detection techniques.  There are several potential ways to defeat such attacks.  First, evaluation of the learned classifier on a validation set or test set may reveal more class confusion than expected (or more than is {\it tolerable}) between classes A and B.
Second, filtering (rejection) of training data based on its {\it provenance} may block ``outsider'' DP attacks.
Third, good system monitoring to detect insider threats is needed to defeat insider DP attacks.
\subsubsection{Data Poisoning in an Unsupervised Learning Context} 

Consider unsupervised clustering/mixture modeling \cite{Duda}, albeit wherein the data batch to be analyzed has been poisoned to include samples from inauthentic clusters/classes.  A good clustering algorithm should identify {\it all} the clusters in the data
batch, including the inauthentic ones.  Moreover, unless the inauthentic clusters are highly unusual\footnote{For example, they may have much greater within-cluster scatter, much greater between-cluster distance to all other clusters than do authentic clusters, or they may have highly atypical feature distributions (e.g., if features are binary, an inauthentic cluster may have an unusually large number of features that are frequently ``on'' (or ``off''), compared with authentic clusters).}, there may be no objective basis for detecting
them as suspicious.  Even if a cluster {\it is} unusual, its distinctiveness can, in fact, alternatively be reasonably interpreted as a {\it validation} of the cluster,
at least compared with other clusters that are {\it too similar} to each other.  The point here is that {\it unsupervised} detection of a DP attack on unsupervised learning is extremely challenging, and possibly ill-posed.

There are two promising approaches to detect such attacks.  One assumes a scenario where there are {\it two} data batches,
one whose content is known to be ``normal'' (representative of a null hypothesis), ${\cal X}_{\rm null}$, and the other, 
${\cal X}_{\rm test}$, which {\it may} contain some anomalous ({\it i.e.}, poisoned) content.  In this setting,
one can formulate a group (cluster) anomaly detection problem, with the null hypothesis that ${\cal X}_{\rm test}$ 
does {\it not} possess content different from ${\cal X}_{\rm null}$ and with the alternative hypothesis that ${\cal X}_{\rm test}$
contains one (or multiple) anomalous clusters relative to ${\cal X}_{\rm null}$.  This problem has been addressed in a quite successful fashion in several
works, even considering the challenging case of huge feature spaces ({\it e.g.}, tens of thousands of features per sample, as seen for document clustering domains), with 
anomalous clusters manifesting on only small, {\it a priori unknown}, subsets of the features \cite{atd,MLSP18-PCAD}.  
These works exploit previous parsimonious mixture modeling techniques \cite{Graham,Hossein}, and treat this hypothesis
testing problem as one of {\it model selection}, with the null hypothesis model (learned on {${\cal X}_{\rm null}$) and 
the alternative model (which consists of the null model plus at least one additional cluster, learned on ${\cal X}_{\rm test}$) competing as the best descriptors
for ${\cal X}_{\rm test}$ in the sense of a model-complexity penalized log-likelihood function such as the 
theoretically grounded Bayesian Information Criterion (BIC) \cite{Schwarz}.  
The alternative hypothesis is chosen when the cost of describing an additional cluster's parameters is more than offset by the associated gain in log-likelihood on the data points that are estimated to belong to this (putative anomalous) cluster. 
These works addressed the case of anomalous clusters in general, {\it i.e.} they did not address DP 
attacks per se, but they are quite suitable to detect them in ${\cal X}_{\rm test}$ when there is also an available (reference)
attack-free batch, ${\cal X}_{\rm null}$. 

When no such reference batch is available, the most promising approach to detect data poisoning is by human operator/analyst
inspection, and perhaps by leveraging information {\it beyond} just the observed data\footnote{Other information sources and knowledge bases may be used by the analyst to suitably ``label'' the clusters to categories pertinent to the application domain.  In this labeling process, anomalous/attack clusters may be discovered.}.

\subsubsection{``Alternative Facts'' Attacks}

We have focused on attacks which induce an ML system to make classification errors.  As shown here, all of these attacks are vulnerable to AD (or clustering based) defenses.  However, there are DP attacks with a different objective which are in general much more formidable to detect.  To illustrate, consider a highly contemporary example in the news.  The Mueller
investigation shared document evidence with a Russian plaintiff's lawyers.  Apparently these lawyers then shared these documents with fake news agents, who altered the documents and then publicly shared them, claiming them as the ``genuine articles''.  This is not an
attack on ML.  It is a data integrity attack \cite{Tygar}, whose goal is to obfuscate what is genuine and factual -- in writ, to create
``alternative facts''.  Such attacks may be for political gain, to alter a criminal trial's outcome, to alter a medical diagnosis
or decision (by altering medical images or test results), and they may even alter or drive a government's policies (image doctoring could lead policy analysts to believe a rogue nation is building WMDs).  Such attacks 
 appear extraordinarily difficult to detect using statistical defenses advocated in this paper -- the facts of a news article can be crucially modified simply by the addition of
one word.  For example: ``Putin did [not] deny that the Russians colluded with Trump...''.  Statistical detection is useless here,
especially since once such fake news is widely promulgated its version of the facts may be just as prevalent (in the news, on 
social networks, and in print) as the genuine facts.  Related attacks on images or audio files, if artfully made, may be similarly resistant to statistical AD.  Again, the key to detecting such attacks may heavily rely on data provenance and {\it e.g.} on
rigorous use of 
cryptographic techniques (as discussed above), including block-chain technology. 

Beyond {\it data} integrity, there are ``entity integrity'' attacks which {\it essentially} seek to pass the Turing test, see {\it e.g.} work on neural dialogue systems \cite{neural-dialogue}.  These attacks are more amenable to statistical AD (or other detection strategies), since increasing the duration and
range of topics covered in a dialogue will eventually test the (common sense skill) limits of an artificial conversant, likely producing responses that are statistically anomalous, logically inconsistent, or which betray large gaps in knowledge of the world. 

\section*{Acknowledgment}
The authors would like to acknowledge research contributions of their student 
Yujia Wang,
which were helpful in the development of this paper.
This research was supported in part by an AFOSR DDDAS grant, a  Cisco Systems
URP gift, and an AWS credits gift.

\bibliographystyle{plain}
\bibliography{bibfiles/computing,bibfiles/ddos,bibfiles/ids,bibfiles/adversarial,bibfiles/kesidis-prior,bibfiles/MyCollection,bibfiles/refs,bibfiles/ref,bibfiles/botsalting,bibfiles/gans,bibfiles/transductive}

\end{document}